\documentclass[acmsmall]{acmart}

\AtBeginDocument{%
  \providecommand\BibTeX{{%
    \normalfont B\kern-0.5em{\scshape i\kern-0.25em b}\kern-0.8em\TeX}}}

\usepackage{amsmath,amsfonts,amsthm}
\usepackage{algorithmic}
\usepackage{graphicx}
\usepackage{textcomp}
\usepackage{xcolor}

\usepackage{amsfonts}       % blackboard math symbols
\usepackage{algorithm}
\usepackage{algorithmic}
\usepackage{bm,url} 
\usepackage{epsfig}
\usepackage{epstopdf}
\usepackage{subfigure} 
\usepackage{comment}
\usepackage{color} 
\usepackage{booktabs}

\newcommand{\MCM}{\mathcal{M}}

\newtheorem{assumption}{{\bf Assumption}}

\newtheorem{definition}{{\bf Definition}}
\newtheorem{example}{{\bf Example}}
\newtheorem{lemma}{{\bf Lemma}}
\newtheorem{problem}{{\bf Problem}}
\newtheorem{theorem}{{\bf Theorem}}

\def\done{\hspace*{\fill} \rule{1.8mm}{2.5mm} \\}

\setcopyright{acmcopyright}
\begin{document}

\title{Analytical and Empirical Study of Herding Effects in Recommendation Systems}

\author{Hong Xie}
\affiliation{%
  \institution{University of Science and Technology of China}  
   \department{State Key Laboratory of Cognitive Intelligence} 
  \country{China}
  }
\email{hongx87@ustc.edu.cn}

\author{Mingze Zhong}
\affiliation{%
  \institution{Chongqing University}  
  \country{China}
  }
\email{zmz@cqu.edu.cn}

\author{Defu Lian}
\affiliation{%
  \institution{University of Science and Technology of China} 
  \department{State Key Laboratory of Cognitive Intelligence}  
  \country{China}
  }
\email{liandefu@cqu.edu.cn}

\author{Zhen Wang}
\affiliation{%
  \institution{Sun Yat-sen University} 
  \country{China}
  }
\email{joneswong.ml@gmail.com}

\author{Enhong Chen}
\affiliation{%
  \institution{University of Science and Technology of China} 
  \department{State Key Laboratory of Cognitive Intelligence}  
  \country{China}
  }
\email{joneswong.ml@gmail.com}

\begin{abstract}
Online rating systems are often used in numerous
web or mobile applications, e.g., Amazon and TripAdvisor,
to assess the \textit{ground-truth quality} of products.
Due to herding effects, the aggregation of historical ratings
(or historical collective opinion) can significantly
influence subsequent ratings,
leading to misleading and erroneous assessments.
We study how to manage product ratings
via rating aggregation rules and
shortlisted representative reviews,
for the purpose of correcting the assessment error.
We first develop a mathematical model to characterize important
factors of herding effects in product ratings.
%,
%and the decision space of an online rating system operator
%in shepherding product ratings.
%
%We identify a class of rating aggregation rules
%as well as sufficient conditions on herding effects
%and review selection mechanisms,
%under which the historical collective opinion
%converges to the ground-truth collective opinion
%of the whole user population.
We then identify sufficient conditions 
(via the stochastic approximation theory),
under which the historical collective opinion
converges to the ground-truth collective opinion
of the whole user population. 
These conditions identify a class of rating aggregation rules
and review selection mechanisms
that can reveal the ground-truth product quality. 
We also quantify the speed of convergence 
(via the martingale theory),
which reflects the efficiency of rating aggregation rules
and review selection mechanisms.
We prove that the herding effects slow down the
speed of convergence while an accurate
review selection mechanism can speed it up. 
We also study the speed of convergence numerically
and reveal trade-offs in selecting rating aggregation rules and
review selection mechanisms. 
To show the utility of our framework, 
we design a maximum likelihood algorithm to infer
model parameters from ratings,
and conduct experiments on rating datasets from Amazon and TripAdvisor.
We show that proper recency aware rating aggregation rules 
can improve the speed of convergence 
in Amazon and TripAdvisor by 41\% and 62\% respectively. 
\end{abstract}


\maketitle

\section{\bf Introduction}

\begin{figure}[htb]
\centering
\subfigure[{\bf Historical collective opinion}]{
\centering
\includegraphics[width=0.38\textwidth]{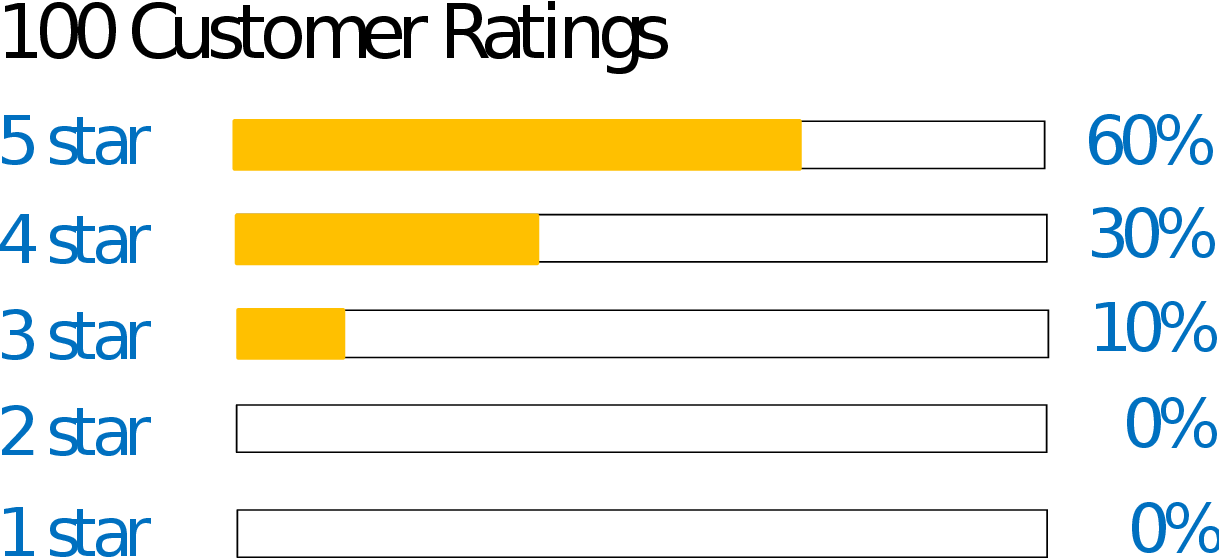}
\label{fig:IntroAmazonAggRat} 
}
\\
\subfigure[{\bf Shortlisted product reviews}]{
\centering
\includegraphics[width=0.38\textwidth]{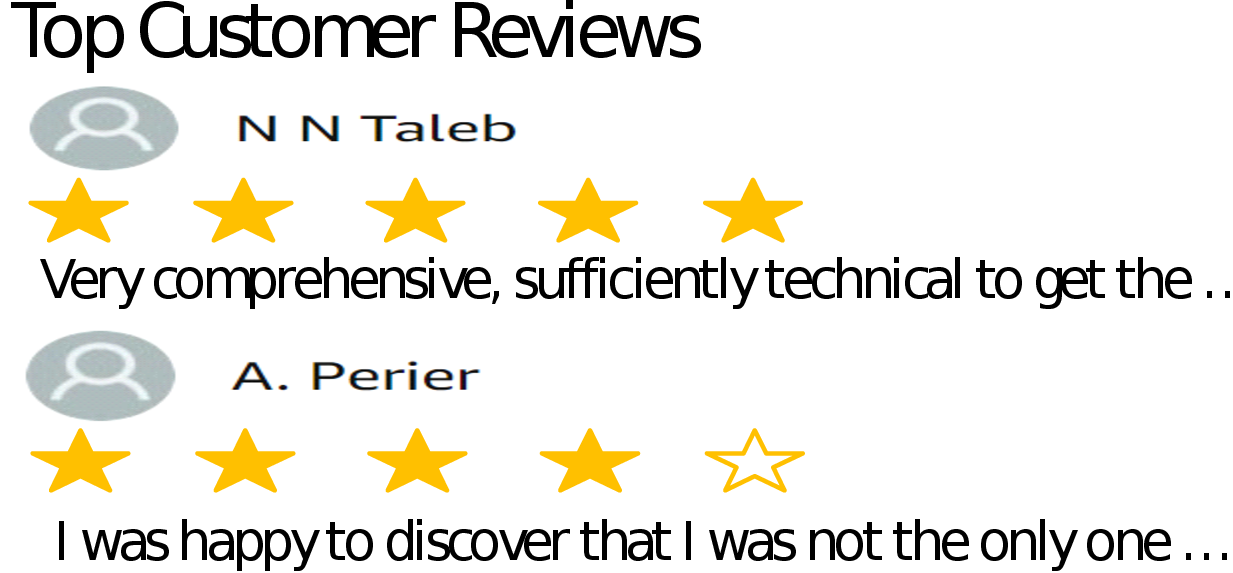}
\label{fig:IntroAmazonRepReview}
}
\caption{(a) Historical collective opinion and
(b) shortlisted product reviews toward a product in Amazon.
}
\label{fig:IntroAmazon}
\end{figure}

Nowadays, online product rating systems are often used
in numerous web or mobile applications,
e.g., Amazon, eBay, TripAdvisor, Google App Store, etc.
Online product rating systems aim to reveal the
\textit{ground-truth quality} of products
via user contributed ratings or reviews.
%For example, such product ratings
%reflect the ground-truth quality of hotels in TripAdvisor.
%, while reflecting ground-truth reputation of sellers in eBay.
Product ratings not only improve users' purchasing experience
\cite{Lackermair2013,Li2013,Zhao2013},
but can also improve revenues of sellers
\cite{Berger2016,Luca2016,Zaroban2015}.
Formally, each user provides ratings to a subset of products,
and their ratings are known to all users.
For each product, the historical collective opinion
(i.e., aggregation of historical ratings) and
shortlisted representative product reviews are usually displayed
to assist users assess the product quality.  
For example, Figure \ref{fig:IntroAmazon}
shows such displays in Amazon.

However, user ratings are ``\textit{biased}'' toward the
displayed historical collective opinion and
shortlisted product reviews due to herding effects
\cite{Muchnik2013,Salganik2006}.
Informally, the herding effects means that users simply ``follow'' the
historical ratings or reviews of the crowd in providing ratings.
This rating bias makes it difficult to reveal
the ground-truth product quality.
To illustrate, let us consider a commonly used rating metric
\{1 = ``Terrible'', 2 = ``Poor'',
3 = ``Average'', 4 = ``Good'', 5 = ``Excellent''\}.
Assume the ground-truth collective opinion of the whole user population is
$(0.01, 0.02, 0.07, 0.4, 0.5)$, i.e., the fraction of users
hold an overall opinion of 1, 2, 3, 4 and 5 are
1\%, 2\%, 7\%, 40\% and 50\% respectively.
Suppose we use the \textit{average scoring rule}
to summarize the collective opinion,
i.e., the \textit{ground-truth quality} is
$0.01\times 1 + 0.02\times 2 + 0.07\times 3 + 0.4\times 4 + 0.5\times 5
\!=\! 4.36$.
For simplicity, 
we use the following two simplified examples to illustrate the
herding effect.

\begin{example}
[{\bf Unbiased ratings}]
\label{exa:Int:HonRev}
Consider the ideal case that users provide unbiased ratings,
i.e., each rating is of 1, 2, 3, 4 and 5, with probability
0.01, 0.02, 0.07, 0.4 and 0.5 respectively.
If the number of historical ratings is sufficiently large \cite{Xie2015},
then its average is around 4.36, which is exactly the
ground-truth quality of the product.
%
%The mean of the historical ratings is
%an accurate estimator of the ground-truth quality
%of the product (i.e., 4.4),
%if the number of historical ratings is
%sufficiently large \cite{Xie2015}.
\end{example}

\begin{example}
[{\bf Ratings under herding effects}]
\label{exa:Int:RevPerCas}
For simplicity, consider one possible herding effect,
i.e., each user provides a rating according
to the historical collective opinion,
i.e., the empirical distribution of past ratings.
Suppose the first rating is 1, then the second rating
will be 1 because the historical collection opinion
is $(1,0,0,0,0)$.
Similarly, the third and all subsequent ratings will be 1.
The average of historical ratings will be 1,
no matter how large the number of ratings is,
and it is very different from the ground-truth quality of 4.36.
\end{example}

Example \ref{exa:Int:HonRev} and \ref{exa:Int:RevPerCas} highlight that
as the strength of herding effects increases,
revealing the ground-truth product quality via historical ratings
(i.e., mean of the historical ratings)
varies from accurate to erroneous.
In general, the strength of herding effects may lie
between that of Example \ref{exa:Int:HonRev} and \ref{exa:Int:RevPerCas},
and the initial ratings is not given in advance.
Some users may even provide high or low ratings intentionally
to promote or badmouth a product.
Furthermore, the shortlisted product reviews
can also influence subsequent ratings.
This paper explores three fundamental questions under such general settings:
\textit{
(1) Under what conditions the historical collective opinion
converges to the ground-truth collective opinion?
(2) What's the convergence speed of the historical collective opinion?
(3) What are some effective rating aggregation rules
and review selection mechanisms to reveal the ground-truth
product quality?
}
The convergence guarantee implies that
the ground-truth product quality can be revealed,
and the rating bias caused by the herding effects can be eliminated.
Namely, it reflects the accuracy of an online rating system.
The speed of convergence reflects
the efficiency of an online rating system,
i.e., a faster speed implies that the ground-truth product quality
can be revealed using a smaller number of ratings.
%We investigate these two questions from the angle of
%shepherding product ratings.
%Given historical ratings,
%different rating aggregation rules, e.g.,
%unweighted aggregation rule, weighted aggregation rule, etc.,
%can produce different historical collective opinion,
%which in turn influences user ratings.
%Furthermore, the shortlisted representative product reviews
%also influences user ratings.
%The operator's decision is to select
%a rating aggregation rule
%and a review selection mechanism
%to influence user ratings, for the purpose
%of eliminating rating bias.
The complicated psychological nature of herding effects
makes it challenging to explore these three questions.
Our contributions are:

\begin{itemize}
\item
We develop a mathematical model to capture important
factors of herding effects in online product ratings.
Our model also characterizes the decision space
of an online rating system operator in selecting rating aggregation rules
and review selection mechanisms.
 
\item
We apply the \textit{stochastic approximation theory}
to derive sufficient conditions,
under which the historical collective opinion
converges to the ground-truth collective opinion
(both the honest and misbehaving rating scenarios).
%This implies that the ground-truth product quality can be revealed. 
These conditions identify a class of rating aggregation rules
and review selection mechanisms,
which can reveal the ground-truth product quality. 

\item
We quantify the speed of convergence
via the ``\textit{martingale theory}'',
which reflects the efficiency of rating aggregation rules
and review selection mechanisms.
We prove that the herding effects slow down the
speed of convergence while an accurate and robust 
review selection mechanism speeds it up. 
We also study the speed of convergence
numerically and find a number of interesting findings.
For example, the improvement of convergence speed
via an accurate selection mechanism becomes
small when we increase the recency awareness
of a rating aggregation rule.  

\item
To show the utility of our framework, 
we design a maximum likelihood algorithm to infer
model parameters from online product ratings 
and conduct experiments on rating datasets from Amazon and TripAdvisor.
We find that TripAdvisor has a higher strength of herding effects
than Amazon 
and appropriate recency aware rating aggregation rules 
can improve the convergence speed in
Amazon and TripAdvisor
by 41\% and 62\% respectively.
\end{itemize}

%5{\color{blue} 
This paper organizes as follows.
Section \ref{sec:model} presents the herding model and the decision model.
Section \ref{sec:analysis} presents the convergence analysis.
Section \ref{sec:Inference} presents a maximum likelihood algorithm
to infer model parameters from data.
Section \ref{sec:exp} and \ref{sec:ExpRealData}
presents the experimental results
on synthetic and real-world data (from Amazon and TripAdvisor)
respectively.
Section \ref{sec:relatedwork} discusses the related work
and Section \ref{sec:conclusion} concludes.
%} 

\section{\bf Model}
\label{sec:model}

We start with the baseline model of unbiased product ratings.
We then model herding effects and
present the decision model in managing product ratings.
Finally, we model misbehavior in ratings.

\subsection{\bf Unbiased Product Rating}

We consider an online product rating system,
which deploys an $M \in \mathbb{N}_{+}$ level cardinal rating metric
to assess product quality
\[
\mathcal{M}
\triangleq
\{
1,\ldots, M
\}.
\]
A higher rating indicates that a user is more satisfied about a product.
For example, a widely deployed rating metric is
$\mathcal{M} =$ \{1 = ``Terrible'', 2 = ``Poor'',
3 = ``Average'', 4 = ``Good'', 5 = ``Excellent''\}.
Without loss of generality,
we consider one product denoted by $\mathcal{P}$.
Note that $\mathcal{P}$ can be interpreted as
a book in Amazon, a hotel in TripAdvisor,
a mobile app in Google App Store, etc.
Let $\alpha_m \in [0,1]$ denote the fraction of users whose
intrinsic (or ground-truth)
opinion toward $\mathcal{P}$ is $m \in \mathcal{M}$.
Namely, the intrinsic (or ground-truth) collective opinion of
the whole user population
toward $\mathcal{P}$ can be characterized by
$
\bm{\alpha} \triangleq [\alpha_1, \ldots, \alpha_M],
$
where $\sum_{m \in \mathcal{M}} \alpha_m =1$.
For example, $M=5$ and
$
\bm{\alpha}
=[0.01, 0.02, 0.07, 0.4 ,0.5]
$
means that 50\% users holds an intrinsic opinion of $5$
toward $\mathcal{P}$.
We denote a space of all the
possible collective opinion vectors as 
\[
\mathcal{O}
\triangleq
\left\{
\bm{\alpha}
\left|
\bm{\alpha} \in [0,1]^M,
\sum_{m \in \MCM} \alpha_m=1
\right.
\right\}.
\] 
 
Let $A: \mathcal{O} \rightarrow [1,M]$ denote an opinion aggregation rule,
which produces an indicator to quantify the
ground-truth quality of a product.
We call $A(\bm{\alpha})$ the ground-truth quality of $\mathcal{P}$.
For example, the commonly deployed
{\em average scoring rule} can be expressed as
$A(\bm{\alpha}) = \sum^M_{m=1} m \alpha_m$.
Furthermore, if
$
\bm{\alpha}
=[0.01, 0.02, 0.07, 0.4, 0.5],
$
we have $A(\bm{\alpha}) = 4.36$.   

Note that $\bm{\alpha}$ is a hidden vector and
online product rating systems aim to reveal it via ratings
provided by users.
Let $R_i \in \mathcal{M}$ denote the $i$-th (in chronological order)
rating of $\mathcal{P}$.
We denote the intrinsic opinion of the user
who provides $R_i$ by $O_i \in \MCM$.
The rating $R_i$ is public to all users,
while the intrinsic opinion $O_i$ is a hidden variable.
We consider a random observation model that $O_i's$
are independent identical distributed (IID) random variables
with a probability mass function (pmf):
\begin{align*}
&
\mathbb{P} [O_i = m ] = \alpha_m,
&&
\forall m \in \MCM,  i \in \mathbb{N}_+.
\end{align*}
%\begin{definition}
We say a rating $R_i$ is ``unbiased'',
if it reflects a user's intrinsic opinion
(i.e., $R_i = O_i$), otherwise it is ``biased''.
%\end{definition}
%Namely, a rating is unbiased only when it reflects a user's intrinsic opinion.
Probabilistically, unbiased ratings
mean that  $R_i's$ are IID random variables with
pmf $\mathbb{P} [R_i = m ] = \alpha_m$.
However, ratings can be biased due to herding effects.
We proceed to model such phenomenon.

\subsection{\bf  Rating Under Herding Effects}

%The historical ratings lead to herding effects in ratings.  
\noindent
{\bf Modeling herding effects. }
We denote all the historical ratings of $\mathcal{P}$
up to the $i$-th rating as
\begin{align*}
&
\mathcal{H}_i
\triangleq
\{R_1, \ldots, R_i \},
&&
\forall i \in \mathbb{N}_+.
\end{align*}
For presentation convenience, we define
$
\mathcal{H}_0 \triangleq \emptyset.
$
We denote historical collective opinion associated with $\mathcal{H}_i$ as
$
\bm{\beta}_i
\!\triangleq\!
[\beta_{i, 1}, \ldots, \beta_{i, M}],
$
where $\beta_{i,m} \in [0,1]$ and
$\sum_{m \in \mathcal{M}} \beta_{i,m} =1$.
Note that the $\bm{\beta}_i$ is public to all users.
Rating recency is important for a variety of applications
\cite{BrightLocal2016,Shrestha2016}.
We consider a \textit{class} of weighted aggregation rules
to capture it, which is expressed as
\begin{align} \label{eq:beta}
&
\beta_{i, m}
\triangleq
\frac{\sum^i_{j=1} w_j \mathbf{I}_{\{R_j = m\}}}
{\sum^i_{j=1} w_j},
&&
\forall m \in \mathcal{M}, i \in \mathbb{N}_+,
\end{align}
where $w_j \geq 0$ denotes the weights for the $j$-th rating,
and $\mathbf{I}$ is an indicator function.
For example, $w_j \!=\! 1, \forall j$, corresponds
to the simple ``\textit{unweighted average rule}'' and
$
\beta_{i, m}
\!=\!
\sum^i_{j=1} \mathbf{I}_{\{R_j = m\}} / i
$
corresponds to the fraction of historical ratings equals $m$.
Thus rule is deployed in many web services
like Amazon, TripAdvisor, etc.
%Namely, $\bm{\beta}_{i}$ is the empirical distribution
%of historical ratings.
Furthermore, $w_i = i$ denotes a
``\textit{recency aware aggregation rule}'',
i.e., assigning higher weights to recent ratings.

Many online rating systems also display
shortlisted representative product reviews.
Each review selection mechanism can be characterized
by the accuracy (in identifying representative reviews)
vs. cost (e.g., complexity) trade-off.
We aim to understand the impact of the selection accuracy
in managing product ratings.
Let
$
\bm{\theta}_i
\!\triangleq\!
\left[
\theta_{i,1}, \ldots, \theta_{i,M}
\right]
$
denote the initial collective opinion
that users form from $\bm{\beta}_i$ and the shortlisted
product reviews,
where $\theta_{i,m} \!\geq\! 0$
and $\sum_{m \in \mathcal{M}} \theta_{i,m} \!=\! 1$.
One interpretation of $\theta_{i,m}$ is
the probability that a user forms an opinion $m \in \mathcal{M}$.
When the review selection mechanism is not deployed,
we model the baseline initial collective
as $\bm{\theta}_i  = \bm{\beta}_i$.

\begin{assumption} \label{asump:ImpRepReview}
Under an accurate review selection mechanism,
it holds that $\|\bm{\theta}_i - \bm{\alpha} \|
\leq
\|\bm{\beta}_i - \bm{\alpha} \|$,
where $\| \cdot \|$ denotes a vector norm.
\end{assumption}

\noindent
Assumption \ref{asump:ImpRepReview} captures
that the initial collective opinion
is closer to the ground-truth collective opinion
than the historical collective opinion,
when representative reviews are presented.

One possible example of $\bm{\theta}_i$ is
\begin{align}\label{example:Thetai}
&
\bm{\theta}_i
= (1 - \eta_i) \bm{\beta}_i + \eta_i \bm{\alpha},
&&
\forall i \in \mathbb{N}_+,
\end{align}
where $\eta_i \in [0,1]$.
Note that $\eta_i = 0$ captures the case that
the review selection mechanism is not deployed,
and $\eta_i$ models the accuracy of
a review selection mechanism
stated in the following lemma.
\begin{lemma}
$||\bm{\theta}_i - \bm{\alpha}||$ is decreasing in $\eta_i$.
\label{lem:ImpEta}
\end{lemma}

\noindent
\textit{ 
Due to page limit, selected proofs are presented in the appendix 
and missing proofs can be found in our supplementary file \cite{Xie2020}. 
}
Lemma \ref{lem:ImpEta} states that
as we increase $\eta_i$,
users form an initial collective opinion $\bm{\theta}_i$
being closer to the ground-truth collective opinion $\bm{\alpha}$.
Namely, increasing the $\eta_i$ models that
the review selection mechanism is more accurate in
selecting representative reviews.
One possible example of $\eta_i$ is
$\eta_i = 0.6 (1 - 1 / i )$,
where $0.6$ captures the accuracy of a
review selection mechanism and
$(1 - 1 / i )$ captures that
the selected reviews is more accurate in reflecting the ground-truth
collective opinion,
when the number of reviews increases.

After purchasing a product, a user's rating is modeled as
a combination of the initial collective opinion and the
intrinsic collective opinion.  Formally, we have
\begin{align*}
&
\mathbb{P} [R_{i} = m | \mathcal{H}_{i-1}]
=\gamma_{i-1} \theta_{i-1,m}
+  (1 - \gamma_{i-1})  \alpha_m,
%&&
%\forall m \in \mathcal{M}, i \in \mathbb{N}_+,
\end{align*}
where $\gamma_i \in [0,1]$ models the strength of herding effects.
Increasing $\gamma_i$ models a stronger strength of herding effects.
We define
$
\bm{\theta}_0 \triangleq \bm{\alpha},
$
to capture that when there is no historical ratings,
a user provides her ground-truth rating.
Note that we consider this simple model for the purpose of
capturing key factors of herding effects while
reducing the number of parameters to tune.
The following assumption eliminates
a trivial case that users purely follow the initial opinion.
\begin{assumption}
The $\gamma_i$
satisfies that $\sup_{i \in \mathbb{N}_+} \gamma_i <1$.
\label{label:Asump:HerdStrength}
\end{assumption}

\noindent
One possible example of $\gamma_i$ is
$\gamma_i  = 0.8 ( 1- 1 / i ) $,
which captures that the strength of herding effects
increases as the number of ratings increases.
Note that our work is general in the sense
that our model is not restricted to
any specific evolving pattern of the strength of herding effects
$\gamma_i$.
In other words, it is allowed
to increase, decrease, or even go up and down in
the number of ratings $i$ (i.e., over time).

\noindent
{\bf Managing product ratings. }
The strength of herding effects $\gamma_i$
is an intrinsic characteristic of the user population,
which the online rating system operators can not control.
To manage product ratings,
their decision is to select the rating aggregation rule, i.e., $w_j$,
and the review selection mechanism, i.e., $\bm{\theta}_i$.
Our objective to make $\bm{\beta}_i$ converge to $\bm{\alpha}$
as fast as possible.
%In other words, reveal the product quality
%(or ground-truth collective opinion) as fast as possible.
This paper aims to provide fundamental understandings
on how to select $w_j$ and $\bm{\theta}_i$.
For the ease of presentation, we denote
$\mathbf{w}
\triangleq
[w_i: i \in \mathbb{N}_+],
\bm{\eta}
\triangleq
[\eta_i: i \in \mathbb{N}_+]$,
$
\bm{\gamma}
\triangleq
[\gamma_i: i \in \mathbb{N}_+]
$
,
and
$
\bm{\Theta}
\triangleq
[\bm{\theta}_i : i \in \mathbb{N}_+].
$

\subsection{\bf Rating Under Misbehavior}

Now, we extend our model to capture misbehaving ratings,
which is also known as spam/fake ratings \cite{Jindal2007}.
It has been reported that some sellers
use fake ratings to promote their own products,
some even use fake ratings to badmouth
their competitors' products \cite{Jindal2007}.
We consider a $(k, \tilde{m}, \mathcal{I})$-misbehavior model,
which is defined as follows.
\begin{definition}
$(k, \tilde{m}, \mathcal{I})$-misbehavior is to inject
$k \in \mathbb{N}_+$ ratings equal to $\tilde{m} \in \mathcal{M}$
toward product $\mathcal{P}$,
where $\mathcal{I} \triangleq \{i_1, \ldots, i_k\}$ denotes
the index set of the injected ratings
and $i_1 < i_2 <\ldots < i_k$.
\end{definition}

For example, a $(2, 5, \{4,5\})$-misbehavior
means injecting two ratings of 5 and the indices of
these two injected ratings are $4, 5$.
Note that this simple misbehavior model
can model many misbehavior
and our objective is to understand
the impact of misbehaving ratings on the
convergence of $\bm{\beta}_i$.

\section{\bf Theoretical Analysis and Implications}
\label{sec:analysis}

We first study the convergence of the
historical collective opinion via the stochastic approximation theory. 
Through this we establish conditions under which
the historical collective opinion converges
to ground-truth collective opinion.   
Then we study the speed of convergence via
the martingale theory.
Through this we identify a metric to guide product rating managing. 
Lastly, we derive the minimum number of ratings to
guarantee an accurate estimation on the
ground-truth product quality.

\subsection{\bf Convergence of Historical Collective Opinion}
 
A commonly used product quality estimation method is $A(\bm{\beta}_i)$.  
Studying the convergence of the historical collective opinion
$\bm{\beta}_i$ is important,
because it lays the foundation for revealing the ground-truth product quality.
In the following theorem, we apply stochastic approximation theory
to investigate the convergence of $\bm{\beta}_i$
under the honest rating scenario, i.e., there are no misbehaving ratings.
\begin{theorem}
Suppose Assumption \ref{asump:ImpRepReview} and
\ref{label:Asump:HerdStrength} hold,
and there are no misbehaving ratings.
If $w_i$ satisfies 
\begin{align}
& \sum^\infty_{i=1} \tilde{w}_i  = \infty,
&&
\sum^\infty_{i=1}
\tilde{w}^2_i
< \infty,
\label{eq:CondAggRule}
\end{align} 
where $\tilde{w}_i \triangleq w_{i} / \sum^{i}_{j=1} w_j$,
then $\bm{\beta}_i$ converges to $\bm{\alpha}$ almost surely, i.e.,
$
\mathbb{P}
\left[
\lim_{i \rightarrow \infty}
\bm{\beta}_i = \bm{\alpha}
\right] = 1.
$
\label{thm:convergence}
\end{theorem}

%\noindent
%{\bf Remark. }
Theorem \ref{thm:convergence} derives sufficient conditions
under which the historical collective opinion $\bm{\beta}_i$
converges to the ground-truth collective opinion $\bm{\alpha}$.
Note that $\bm{\beta}_i$ converges to $\bm{\alpha}$ implies
that the ground-truth quality will be revealed,
i.e., $A(\bm{\beta}_i)$ converges to $A(\bm{\alpha})$.
In other words, the rating bias caused by herding effects
will eventually be eliminated.
It is important to note that the convergence of $\bm{\beta}_i$
is achieved without adding any condition on $\eta_i$.
This means that the convergence can be achieved even
without any review selection mechanism.
Condition (\ref{eq:CondAggRule}) identifies a class of
weighted aggregation rules to guarantee the convergence
of historical collective opinion.
It characterizes a broad class
of aggregation rules for the online rating system operator to choose
as we proceed to illustrate.

Condition (\ref{eq:CondAggRule}) characterizes a
large supply of rating aggregation rules.
Let us use some examples to illustrate this point.
Consider $w_i =1, \forall i \in \mathbb{N}_+$,
which corresponds to the simple unweighted average rule.
We have $\tilde{w}_i = w_i / \sum^i_{j=1} w_j = 1/i$.
One can easily check that Condition (\ref{eq:CondAggRule}) holds,
i.e.,
\begin{align*}
&
\sum^\infty_{i=1}
\tilde{w}^2_i =
\sum^\infty_{i=1}
\frac{1}{i}
= \lim_{i \rightarrow \infty } \ln i=\infty,
&&
\sum^\infty_{i=1}
\tilde{w}^2_i
=
\sum^\infty_{i=1}
\frac{1}{i^2}
<2.
\end{align*}
Namely, the simple unweighted average rule is a candidate. 
Consider a recency aware aggregation rule with $w_i =i$.
Under this aggregation rule, recent ratings are of higher importance.
We have
$\tilde{w}_i = i / \sum^i_{j=1} j = 2/(i+1)$.
One can easily check that Condition (\ref{eq:CondAggRule}) also holds.
Namely, for applications that the recency of rating matters,
the online rating system operator can choose this aggregation rule.
Consider a more general example $w_i = i^c, \forall c\geq 0$.
First, one can derive $\sum^i_{j=1} w_j$ as
\[
\sum^i_{j=1} w_j
= \sum^i_{j=1}  j^c
\approx
\int^i_0 x^c dx
= \left. \frac{x^{c+1} }{c +1} \right|^i_0
= \frac{i^{c+1} } {c +1}.
\]
Then, one can calculate $\tilde{w}_i$ as
$
\tilde{w}_{i}
\!\approx\! (c+1) / i.
$
We conclude that this class of
recency aware aggregation rules also
satisfy Condition (\ref{eq:CondAggRule}).
%Namely, we have a large supply of rating aggregation
%rules to choose. 

In the following theorem,
we extend Theorem \ref{thm:convergence} to study the
impact of misbehaving ratings.

\begin{theorem}
Suppose Assumption \ref{asump:ImpRepReview},
\ref{label:Asump:HerdStrength},
and
Condition (\ref{eq:CondAggRule}) hold.
If the $(k, \tilde{m}, \mathcal{I})$-misbehavior
satisfies $k < \infty$,
then $\bm{\beta}_i$ converges to $\bm{\alpha}$ almost surely, i.e.,
$
\mathbb{P}
\left[
\lim_{i \rightarrow \infty}
\bm{\beta}_i = \bm{\alpha}
\right] = 1.
$
\label{thm:ImpMisbConv}
\end{theorem}

Theorem \ref{thm:ImpMisbConv} states that the convergence of the
historical collective opinion is invariant of the misbehaving ratings,
as long as the number of misbehaving ratings is finite.
It states that the aggregation rules that satisfy
Condition (\ref{eq:CondAggRule}) are
\textit{robust} against misbehaving rating attacks.  
This implies that the online rating system operator
does need to worry about misbehaving ratings,
as long as its number of misbehaving ratings is finite.
%the ground-truth product quality can be revealed. 

\subsection{\bf Convergence Speed of Historical Collective Opinions}

The convergence speed of historical collective opinion reflects the
efficiency of online rating systems,
because a faster speed implies that the ground-truth product quality
can be revealed with a smaller number of ratings.
Under general initial opinion vector $\bm{\theta}_i$,
it is difficult to study the convergence speed of
$\bm{\beta}_i$ analytically.
We therefore we focus on one class of initial opinion vector
$\bm{\theta}_i$ derived in Equation (\ref{example:Thetai}).
The convergence speed for this case
can already provide important insights on
selecting rating aggregation rules and review selection mechanisms.
In the following theorem, we apply martingale theory to
study the honest rating scenario.

\begin{theorem}
Suppose Assumption \ref{asump:ImpRepReview} and
\ref{label:Asump:HerdStrength} hold,
and there are no misbehaving ratings.
Suppose $w_i$ satisfies
Condition (\ref{eq:CondAggRule})
and $\bm{\theta}_i$ satisfies Equation (\ref{example:Thetai}).
Let $\epsilon \in [0,1]$ denote an estimation error.
For each $m \in \mathcal{M}$ and $i \in \mathbb{N}_+$, we have
\begin{align*}
&
\mathbb{P}
\left[ |\beta_{i,m} - \alpha_m| > \epsilon \right]
\leq
2 \exp \left( -\phi_i (\mathbf{w}, \bm{\eta}, \bm{\gamma}) \epsilon^2 \right),
%&&
%\forall i \in \mathbb{N}_+,
\end{align*}
where $\phi_i  (\mathbf{w}, \bm{\eta}, \bm{\gamma})$ is defined as
\begin{align*}
&
\phi_i (\mathbf{w}, \bm{\eta}, \bm{\gamma})
\triangleq
\frac{1}
{
\frac{\varphi^2_{i-1} (\mathbf{w}, \bm{\eta}, \bm{\gamma}) }{2}
\sum\nolimits^i_{j=1}
\frac{\tilde{w}^2_j }
{\varphi^2_{j-1} (\mathbf{w}, \bm{\eta}, \bm{\gamma})}
},
&&
\forall i \in \mathbb{N}_+,
\\
&
\varphi_j (\mathbf{w}, \bm{\eta}, \bm{\gamma})
\triangleq
\prod^j_{\ell=1}
\left[
1 - \tilde{w}_{\ell+1} (1 - \gamma_{\ell} + \eta_{\ell}  \gamma_{\ell} )
\right],
&&
\forall j \in \mathbb{N}_+,
\end{align*}
and we define
$\varphi_0 (\mathbf{w}, \bm{\eta}, \bm{\gamma}) \triangleq 1$.
\label{thm:ConvRate}
\end{theorem}

Theorem \ref{thm:ConvRate} derive a metric, i.e.,
$\phi_i(\mathbf{w}, \bm{\eta}, \bm{\gamma})$,
to quantify the convergence speed of the historical collective opinion.
Given the number of ratings $i$,
larger $\phi_i(\mathbf{w}, \bm{\eta}, \bm{\gamma})$
implies faster convergence speed.
The $\phi_i(\mathbf{w}, \bm{\eta}, \bm{\gamma})$
serves as a building block to study product rating managing.
It enables us to analyze the impact of
$\mathbf{w}, \bm{\eta}, \bm{\gamma}$ on the convergence speed,
so as to draw important insights on managing product ratings.
Algorithmically, it enables online rating system operators
to design algorithms to select proper rating aggregation rule
and review selection mechanisms to speed up convergence.
%
%Furthermore, $\phi_i(\mathbf{w}, \bm{\eta}, \bm{\gamma})$
%can also serves as a metric to measure the efficiency of
%a rating aggregation rule (i.e., $\mathbf{w}$) and
%review selection mechanism (i.e., $\bm{\eta}$).  %
%An online rating system operator can select
%appropriate $\mathbf{w}, \bm{\eta}$ to
%increase the convergence speed.
%Lastly, $\phi_i(\mathbf{w}, \bm{\eta}, \bm{\gamma})$ enables
%us to understand the impact of strength of herding effects
%(i.e., $\bm{\gamma}$) on the convergence speed.

To illustrate, let us consider the simple unweighted average rule
with $w_i =1$
and $\gamma_i = 0$ (i.e., there is no herding bias).
In this special case,
ratings are independently and identically generated
according to $\bm{\alpha}$.
We have $\tilde{w}_{\ell} = 1/\ell$,
$\varphi_j (\mathbf{w}, \bm{\eta}, \bm{\gamma})
= \prod^j_{\ell=1} (1 - 1/(\ell+1)) = 1 / (j+1)$.
Then it follows that
$\phi_i (\mathbf{w}, \bm{\eta}, \bm{\gamma})
= 1 / (\frac{1}{2 i^2} \sum^i_{j=1} 1 ) = 2 i$.
Note that this corresponds to the Chernoff bound for IID ratings.
This example shows that the Chernoff bound is a special case
of Theorem \ref{thm:ConvRate}.
In the following theorem, we characterize the impact of
$ \bm{\eta}$ and $\bm{\gamma}$ on the speed of convergence.

\begin{theorem}
$\phi_i (\mathbf{w}, \bm{\eta}, \bm{\gamma})$
is non-increasing in $\gamma_{j}, \forall j \leq i$,
and non-decreasing in $\eta_{j}, \forall j \leq i$.
\label{thm:ImpGamEtaSpeed}
\end{theorem}

Theorem \ref{thm:ImpGamEtaSpeed} states that
$\phi_i (\mathbf{w}, \bm{\eta}, \bm{\gamma})$
decreases as the strength of herding effects $\gamma_{j}$ increases,
and increases as the accuracy of the review selection mechanism
$\eta_j$ increases.
This implies that as users' ratings are more prone to herding effects,
the speed convergence slows down ,
and it speeds up if the online rating system operator
can improve the accuracy of the review selection mechanism.
The impact of aggregation rules (i.e., $\mathbf{w}$)
is not as clear as $\bm{\eta}$ and $\bm{\gamma}$,
and we will study it through numerical analysis. 
In the following theorem, we study the impact of
misbehaving ratings.

\begin{theorem}
Suppose Assumption \ref{asump:ImpRepReview}
and \ref{label:Asump:HerdStrength} hold.
Suppose Condition (\ref{eq:CondAggRule}) hold,
and $\bm{\theta}_i$ satisfies Equation (\ref{example:Thetai}).
If the $(k, \tilde{m}, \mathcal{I})$-misbehavior
satisfies $k < \infty$,
%Let $\epsilon \in [0,1]$ denote the estimation error.
then we have
\begin{align*}
&
\mathbb{P} [|\beta_{i,m} - \alpha_m| > \epsilon]
\leq
2 \exp\left(
- \tilde{\phi}_i (\mathbf{w}, \bm{\eta}, \bm{\gamma})
\epsilon^2
\right),
%&&
%\forall i > i_k,
\end{align*}
where $\tilde{\phi}_i  (\mathbf{w}, \bm{\eta}, \bm{\gamma})$ is defined as
\begin{align*}
\tilde{\phi}_i (\mathbf{w}, \bm{\eta}, \bm{\gamma})
\triangleq
&
\mathbf{I}_{
\big\{
\frac{ \varphi_{i-1} (\mathbf{w}, \bm{\eta}, \bm{\gamma}) }
  {\epsilon \varphi_{i_k -1}(\mathbf{w}, \bm{\eta}, \bm{\gamma})}
\leq 1
\big\}
}
\left(
1 -
\frac{ \varphi_{i-1} (\mathbf{w}, \bm{\eta}, \bm{\gamma}) }
  {\epsilon \varphi_{i_k -1}(\mathbf{w}, \bm{\eta}, \bm{\gamma})}
\right)^2
\\
&
\bigg/
\left(
\frac{\varphi^2_{i-1} (\mathbf{w}, \bm{\eta}, \bm{\gamma})}{2}
\sum^i_{j=i_k + 1}
\frac{\tilde{w}^2_{j}}
  {\varphi^2_{j-1} (\mathbf{w}, \bm{\eta}, \bm{\gamma})}
\right)
\end{align*}
for all $i > i_k$,
$\tilde{\phi}_i (\mathbf{w}, \bm{\eta}, \bm{\gamma})
\triangleq 0$ for all $i \leq i_k$
and $\varphi_{-1} (\mathbf{w}, \bm{\eta}, \bm{\gamma})
\triangleq \infty$.
Furthermore, $\tilde{\phi}_i (\mathbf{w}, \bm{\eta}, \bm{\gamma}) \leq
\phi_i (\mathbf{w}, \bm{\eta}, \bm{\gamma})$
for all $k \geq 1$
and
$\tilde{\phi}_i (\mathbf{w}, \bm{\eta}, \bm{\gamma}) =
\phi_i (\mathbf{w}, \bm{\eta}, \bm{\gamma})$ for $k=0$ and $i_0 = 0$.
\label{thm:ImpMisbConvRate}
\end{theorem}

Theorem \ref{thm:ImpMisbConvRate} derives a metric,
i.e., $\tilde{\phi}_i (\mathbf{w}, \bm{\eta}, \bm{\gamma})$,
to quantify the convergence speed of historical collective opinion
under misbehaving rating attacks.
It states that misbehaving ratings slows down the convergence,
as $\tilde{\phi}_i (\mathbf{w}, \bm{\eta}, \bm{\gamma}) \leq
\phi_i (\mathbf{w}, \bm{\eta}, \bm{\gamma})$.
Theorem \ref{thm:ConvRate} is a special case
of Theorem \ref{thm:ImpMisbConvRate} with $k=0$
and $i_0 = 0$. 
 
\subsection{\bf  Product Quality Estimation}

As an application of the above convergence rate metric,
we now derive the minimum number of ratings needed to
reveal the intrinsic product quality.
In particular, we consider two commonly used
opinion aggregation rules:
\begin{align}
&
\text{average scoring rule: }
A(\bm{\alpha})
= \sum^M_{m=1} m \alpha_m,
\label{eq:asr}
\\
&
\text{majority rule: }
A(\bm{\alpha})
= \arg\max_{m \in \mathcal{M}} \alpha_m.
\label{eq:mjr}
\end{align}
Note that these two opinion aggregation rules are not conflicting.
They reflect two perspectives on producing an indicator
on the product quality.
For brevity, we only consider the honest rating case,
and one can easily extend to accommodate misbehaving ratings
based on Theorem \ref{thm:ImpMisbConvRate}.
In the following theorem, we apply theorem \ref{thm:ConvRate}
to derive the minimum number of ratings needed.
\begin{theorem}
Suppose the same assumptions in Theorem \ref{thm:ConvRate} hold.
Consider the average scoring rule, i.e., $A$ satisfies (\ref{eq:asr}).
If the number of ratings satisfies
\[
i \geq
\phi^{-1}
\left(
\frac{M^2 (M+1)^2}{4 \epsilon^2}  \ln \frac{2M}{ \delta}
\right),
\]
then $|A(\bm{\beta}_i) - A(\bm{\alpha})| \leq \epsilon$
holds with probability at least $1 - \delta$,
where the function $\phi^{-1}$ is defined as
\[
\phi^{-1} (x)
\triangleq
\arg \max_{i \in \mathbb{N}_+}
\{ \phi_i (\mathbf{w}, \bm{\eta}, \bm{\gamma}) \leq x \}.
\]
Consider the majority rule, i.e., $A$ satisfies (\ref{eq:mjr}).
If the number of ratings satisfies
\[
i \geq
\phi^{-1}
\left(
\frac{M^2 (M+1)^2}{ (\alpha_{max} - \alpha_{secmax})^2}
\ln \frac{2 M}{\delta}
\right),
\]
then $A(\bm{\beta}_i) = A(\bm{\alpha})$
holds with probability at least $1 - \delta$,
where $\alpha_{max}$ and $\alpha_{secmax}$
denotes the largest and second largest element
of the vector $\bm{\alpha}$ respectively.
\label{cor:minNumRat}
\end{theorem}
Theorem \ref{cor:minNumRat} derives lower bounds on the number of ratings
such that the product estimation is accurate.
It states that the number of ratings needed is critical to the accuracy
$\epsilon$ for the average scoring rule,
and is critical to the opinion gap $|\alpha_{max} - \alpha_{secmax}|$
for the majority rule.
For other opinion aggregation rules such as the median rating rule,
one can apply Theorem \ref{thm:ConvRate} to derive similar bounds
on the number of ratings needed.
 
\section{\bf Inferring Model Parameters}
\label{sec:Inference}
 
In this section, we first present a maximum likelihood estimation
(MLE) framework to infer model parameters from ratings.
We show that without adding extra conditions,
there is an issue of over fitting caused by the
high dimensionality of model parameters.
To resolve this issue, we propose a linear approximation approach.  
 
\subsection{\bf MLE Framework \& the Over Fitting Issue}
 
\noindent
{\bf MLE framework. } 
%\begin{Conference}
%{\bf MLE framework \& the over fitting issue. }
%\end{Conference}
Without loss of generality, we focus one infer
the model parameters for one product.
We consider the scenario that we are given
$N \in \mathbb{N}_+$ ratings $R_1, \ldots, R_N$
of product $\mathcal{P}$
and the associated rating aggregation rule $\mathbf{w}$.
Our objective is to infer $\bm{\alpha}, \bm{\gamma}$ and
$\bm{\Theta} = [\bm{\theta}_i : i \in \mathbb{N}_+] $
from these ratings via maximum likelihood estimation.
The log-likelihood function can be derived as
\begin{align*}
L(\bm{\alpha}, \bm{\gamma}, \bm{\Theta})
%& \triangleq \ln \MCL (\bm{\alpha}, \bm{\beta}, \gamma)
%\\
%&
=
\sum^N_{i=2}
\ln
\left(\gamma_{i-1} \theta_{i-1,R_i}
+  (1 - \gamma_{i-1})  \alpha_{R_i}
\right).
\end{align*}
%Formally, we state our inference problem as follows.

\begin{problem} \label{prob:Inference:Opt}
Given $R_1, \ldots, R_N$ of the product $\mathcal{P}$ and $\mathbf{w}$.
Select model parameters to maximize
$
L(\bm{\alpha}, \bm{\gamma}, \bm{\Theta})
$:
\begin{align}
&
\underset{\bm{\alpha}, \bm{\gamma}, \bm{\Theta}}{\text{maximize}}
&&
L(\bm{\alpha}, \bm{\gamma}, \bm{\Theta})
&&
\nonumber
\\
& \text{subject to} &&
\|\bm{\theta}_i - \bm{\alpha} \|
\leq
\|\bm{\beta}_i - \bm{\alpha} \|, \,\,\,
\bm{\beta}_i \text{ satisfies Eq. (\ref{eq:beta})},
\nonumber
\\
&
&& \sum_{m \in \mathcal{M}} \alpha_m = 1,
\hspace{0.18 in}
\sum_{m \in \MCM} \theta_{i,m} =1,
\nonumber
\\
& && \bm{\alpha} \in [0,1]^{M}, \,\,\,
\bm{\theta}_i \in [0,1]^{M}, \,\,\,
\gamma_i \in [0,1].
\nonumber
\end{align}
\end{problem}
 
\noindent
{\bf The over fitting issue. } 
The following theorem characterizes the optimal solution of
Problem \ref{prob:Inference:Opt},
which reveals an issue of over fitting.

\begin{theorem} \label{thm:InfExistence}
The optimal solution $\bm{\alpha}^\ast, \bm{\Theta}^\ast$
and $\bm{\gamma}^\ast$ of
Problem \ref{prob:Inference:Opt} satisfies
\[
\gamma^\ast_i
=
\left\{
\begin{aligned}
&
0,
&&
\text{ if $\theta^\ast_{i,R_{i+1}} \leq \alpha^\ast_{R_{i+1}}$},
\\
&
1,
&&
\text{ otherwise}.
\end{aligned}
\right.
\]
where $i \leq N-1$.
Furthermore, $\bm{\theta}^\ast_i$ and $\gamma^\ast_i$ are arbitrary
for all $i \geq N$.
\end{theorem}

\noindent
Theorem \ref{thm:InfExistence} states
that the inferred strength
of herding effects $\gamma^\ast_i$ is either
0 or 1 for all $i\leq N-1$,
and is arbitrary for all $i \geq N$.
This statement holds no matter what ratings are fed to
the Problem \ref{prob:Inference:Opt}.
It implies an issue of over fitting
and the inferred strength of herding effects is not meaningful.
The inferred initial opinion $\bm{\Theta}^\ast$
also has similar over fitting issues
and we omit it for brevity.
One reason for the over fitting is that the number of model parameters
is far more than the number of ratings.
We next propose a linear approximation approach to reduce
the dimension of the model parameters,
while making the inferred parameters
meaningful and interpretable.

%\begin{Journal}
\subsection{\bf Linear Approximation}
\label{subsec:LinApprox}
%\end{Journal}
%\begin{Conference}
%\noindent 
%{\bf Linear approximation. }
%\end{Conference}
To address the over fitting issue,
we consider the linear approximation of $\bm{\theta}_i$
derived in Equation (\ref{example:Thetai}),
where the parameter $\eta_i$ is interpreted as
the accuracy of a review selection mechanism.
To further reduce the number of parameters to tune,
we set $\eta_i = \eta, \gamma_i = \gamma,
\forall i \in \mathbb{N}_+$.
The $\eta$ and $\gamma$ can be interpreted as the
overall accuracy
of a review selection mechanism or overall strength of herding effects.
Then, it boils down to infer $\bm{\alpha}, \eta, \gamma$ from ratings.
Log-likelihood function is 
\[
L(\bm{\alpha}, \eta, \gamma)
\!=\!
\prod^N_{i=2}
\ln
\left[  (1-\eta) \gamma \beta_{i-1, R_i}
+ (1 -  (1-\eta) \gamma) \alpha_{R_i}
\right].
\]
One can observe that $\eta$ and $\gamma$ is not identifiable.
This is because two pairs
$(\eta, \gamma)$ and $(\eta', \gamma')$
can give the same log-likelihood, i.e.,
$
L(\bm{\alpha}, \eta, \gamma)
=L(\bm{\alpha}, \eta', \gamma')
$,
if $(1-\eta) \gamma = (1-\eta') \gamma'$.
To resolve this issue, we
define $\tilde{\gamma} \triangleq (1-\eta) \gamma$,
which can be interpreted as effective strength of herding effects
under a review selection mechanism.
The corresponding log-likelihood function is 
\[
L(\bm{\alpha}, \tilde{\gamma})
=\prod^N_{i=2}  \ln
\left[  \tilde{\gamma} \beta_{i-1, R_i}
+ (1 -  \tilde{\gamma}) \alpha_{R_i}
\right].
\]

\noindent
The inference problem can be stated as follows.

\begin{problem} \label{prob:Inference:LinApprox}
Given $R_1, \ldots, R_N$ of the product $\mathcal{P}$ and $\mathbf{w}$.
Select $\bm{\alpha}, \tilde{\gamma}$ to maximize
$
L(\bm{\alpha}, \tilde{\gamma})
$:
\begin{align}
&
\underset{\bm{\alpha}, \tilde{\gamma}}{\text{maximize}}
&&
L(\bm{\alpha}, \tilde{\gamma})
&&
\nonumber \\
& \text{subject to}
&&
\sum_{m \in \mathcal{M}} \alpha_m = 1,
\hspace{0.18 in}
 \bm{\beta}_i \text{ satisfies Eq. (\ref{eq:beta})},
\nonumber
\\
& && \bm{\alpha} \in [0,1]^{M},
\hspace{0.18 in}
\tilde{\gamma} \in [0,1].
\nonumber
\end{align}
\end{problem}

Problem \ref{prob:Inference:LinApprox} reduces the number of
models parameters from infinite to $M+1$.
Through this we resolve the over fitting issue of Problem
\ref{prob:Inference:Opt}, and all the $M+1$
parameters have clear physical meaning.
Note that the objective function of
Problem \ref{prob:Inference:LinApprox}
is non-linear and not concave.
This means that it may have multiple
local optimal solutions.
One can apply a gradient method to locate one local
optimal solution of it.
To increase the chance of hitting one global optimal point,
one can repeat the gradient method with multiple
different initial points.
Through this we may obtain multiple local optimal solutions,
and among them we select the one with the largest
objective functional value.
We denote the selected one as
$\bm{\alpha}_N, \tilde{\gamma}_N$. 
In the next section, we evaluate the
accuracy of this search scheme via
experiments on synthetic data.

\section{\bf Experiments on Synthetic Data}
\label{sec:exp}

We conduct experiments on synthetic data to quantitatively
study the impact of $\mathbf{w}, \bm{\eta}$ and $\bm{\gamma}$
on the convergence speed.
We also evaluate the accuracy of our inference algorithm.  
\textit{ 
Code and dataset can be found in the link\footnote{\url{https://1drv.ms/u/s!AkqQNKuLPUbEii1kw6asVjdSKXUm?e=y9c8Nl}}.  
}

\subsection{\bf Evaluating the Convergence Speed}
\label{subsec:expSynConSp}
%Due to page limit, we focus on the honest rating case,
%i.e., there are no misbehaving ratings,
%which can already reveal fundamental understandings.
 
\begin{figure*}[htb] 
\centering
\subfigure[Herding strength $\gamma = 0$]{
\centering
\includegraphics[width=0.3\textwidth]{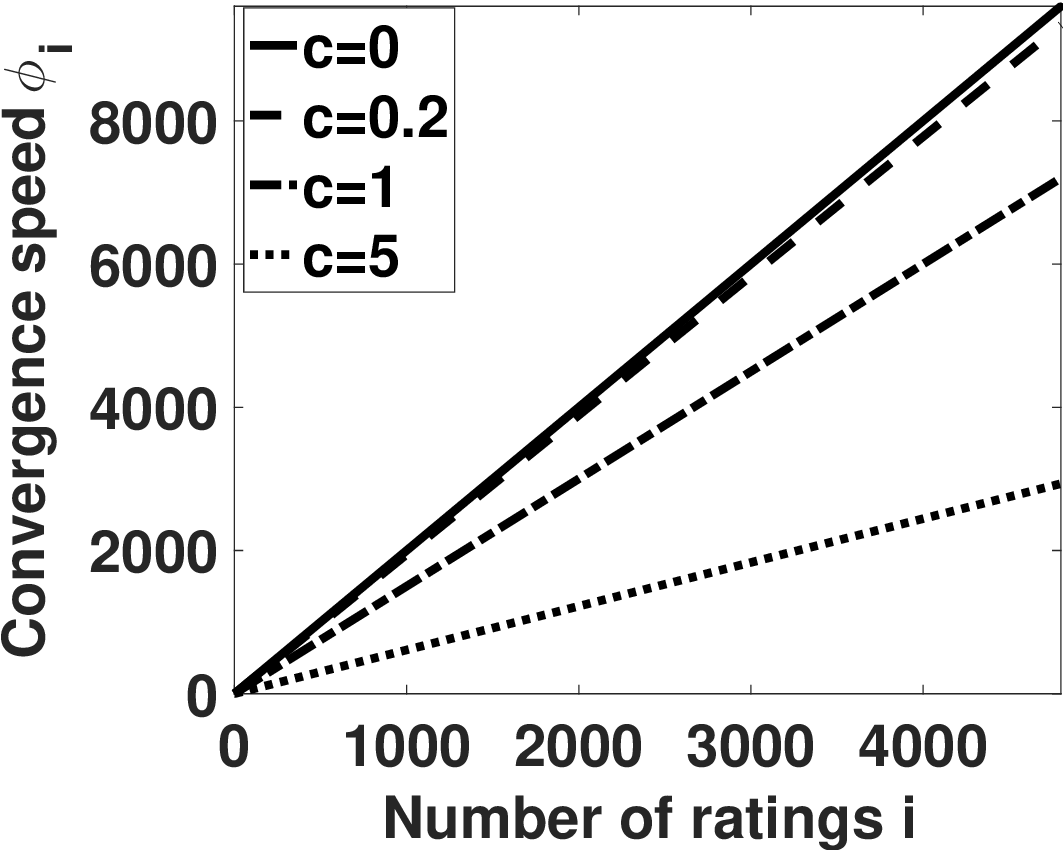}
\label{fig:ImpcSpeedConGam0}
} 
\subfigure[Herding strength $\gamma = 0.2$]{
\centering
\includegraphics[width=0.3\textwidth]{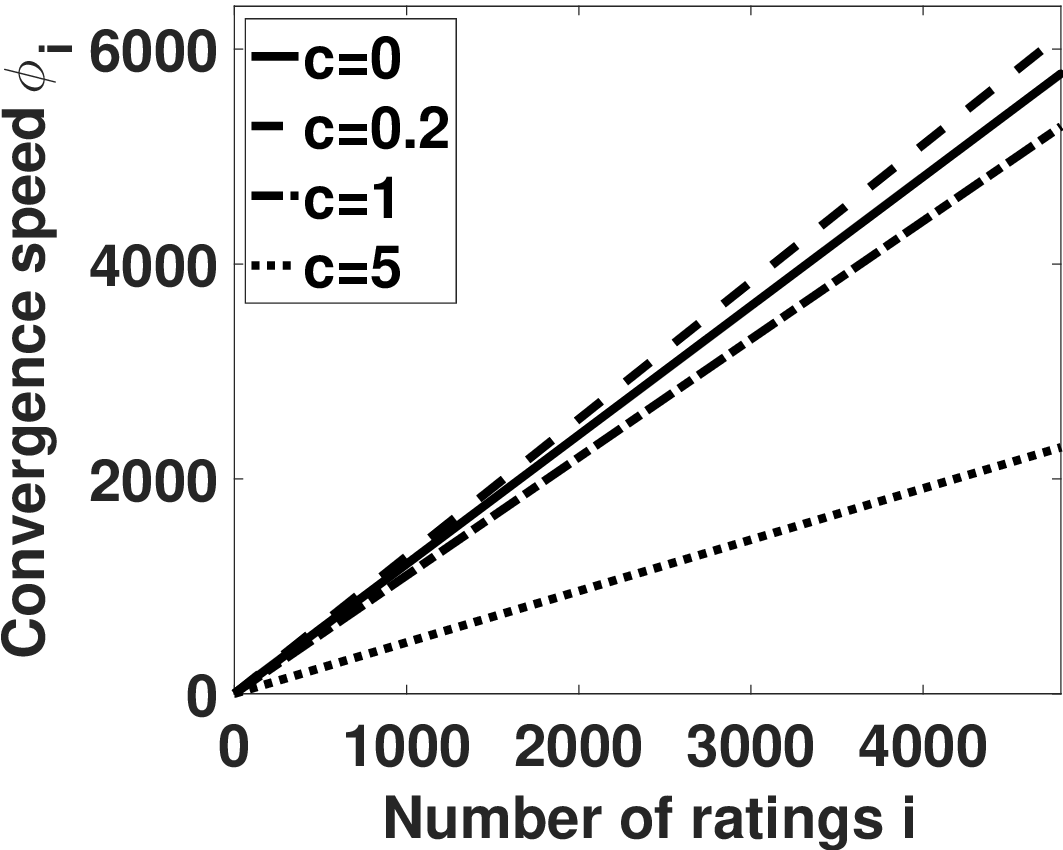}
\label{fig:ImpcSpeedConGam02}
} 
\subfigure[Herding strength $\gamma = 0.4$]{
\centering
\includegraphics[width=0.3\textwidth]{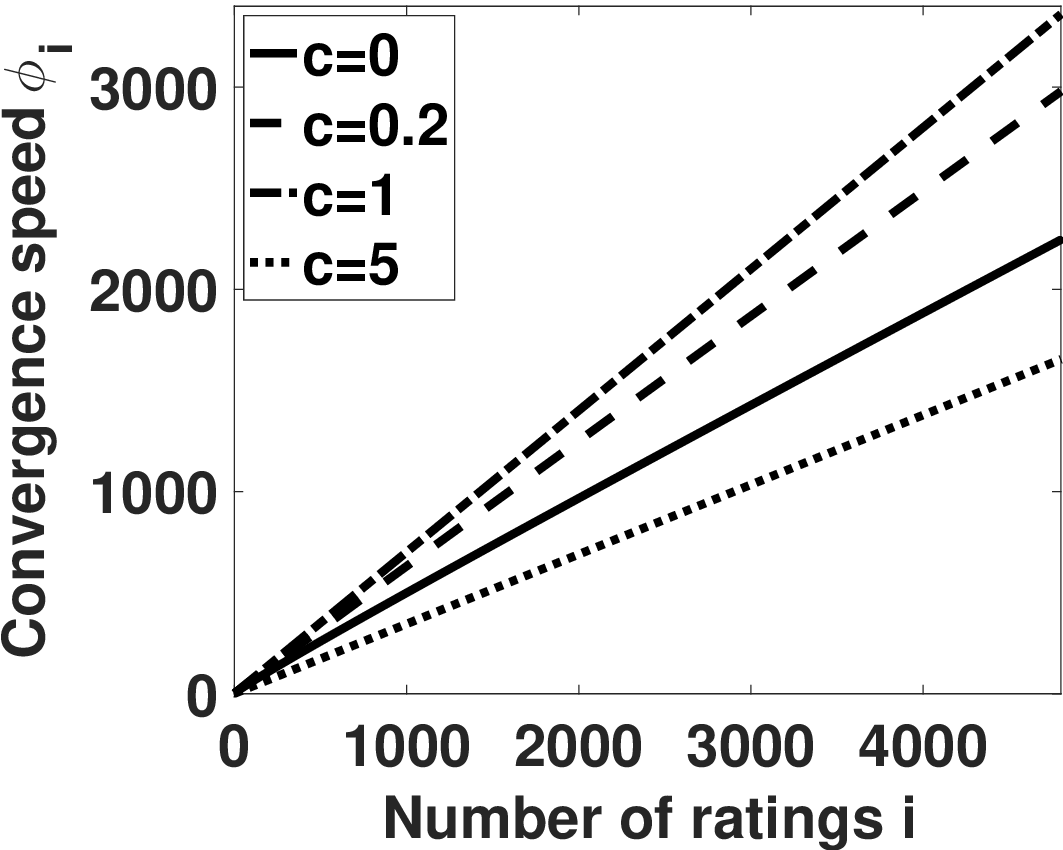}
\label{fig:ImpcSpeedConGam04}
}
\caption{Impact of rating aggregation rules (i.e., $c$) on the speed of convergence, where $\eta =0$
and $\phi_i$ denotes
$\phi_i (\mathbf{w}, \bm{\eta}, \bm{\gamma})$.  }
\label{fig:ImpCSpeedCon}
\end{figure*}

We focus on the honest rating scenario,
because the convergence speed under misbehaving ratings
has similar analytical expressions as the honest rating case.
The importance of rating recency was justified by
a variety of evidences in real-world
applications \cite{BrightLocal2016,Shrestha2016}.
To capture it, we consider
a class of rating aggregation rules with
\[
w_i = i^c,
\]
where $c \in [0, \infty)$.
Recall from Section \ref{sec:analysis} that
this class of rating aggregation rules can reveal
the ground-truth collective opinion.
The parameter $c$ models the strength of the recency awareness.
Increasing $c$ means that the rating aggregation rule is
more aware of the recency,
and $c = 0$ captures that the recency is not considered.
To study the impact of the accuracy of review selection mechanisms,
we consider the initial collective opinion $\bm{\theta}_i$
derived in Equation (\ref{example:Thetai}),
i.e., $\eta_i$ models the accuracy.
For the purpose of reducing the number of parameters to tune,
we set $\eta_i = \eta$
and $\gamma_i = \gamma$.
The $\eta$ and $\gamma$ can be interpreted as the overall accuracy
of a review selection mechanism and overall strength of herding effects
respectively.
%We aim to vary $c, \eta$ and $\gamma$ to study their impact individually.

\noindent
{\bf Impact of rating aggregation rules ($\bm{w}$).}
Recall that $w_i  = i^c$.
Thus, we study the impact of
rating aggregation rules $\mathbf{w}$ through varying $c$.
Figure \ref{fig:ImpCSpeedCon} shows the curve of
$\phi_i (\mathbf{w}, \bm{\eta}, \bm{\gamma})$
across $i$, where we vary $c$ from 0 to 5.
Note that in the figure, we use $\phi_i$ to denote
$\phi_i (\mathbf{w}, \bm{\eta}, \bm{\gamma})$ for brevity.
From Figure \ref{fig:ImpCSpeedCon}, one can observe that
the value of $\phi_i$ is almost linear in the number of ratings $i$.
This implies that the speed of convergence is roughly exponential in the
number of ratings $i$.
Figure \ref{fig:ImpcSpeedConGam0} shows that $\phi_i$ decreases
in $c$ when $\gamma=0$.
%In other words, when the strength of herding is zero,
%the speed of convergence decreases in $c$.
This implies that when there are no herding effects,
the convergence speed of
the simple unweighted average rule is faster
than recency aware aggregation rules.
However, when $\gamma=0.2$ and $\gamma=0.4$,
$\phi_i$ first increases and then decreases as we increase $c$.
This implies that in the presence of herding effects,
recency aware aggregation rule can have faster speed of convergence.
However, the strength of recency awareness should not be too strong.
Furthermore, when $\gamma=0.2$ and $\gamma=0.4$,
the aggregation rules with $c=0.2$ and $c=1$
have the highest convergence speed respectively.
Namely, as the strength of herding effects increases,
the online rating system operator can increase the
strength of recency awareness to speed up convergence.

%{\bf Lessons learned. }
%The speed of convergence is exponential in the number of ratings.
%The aggregation rule that assign higher weights
%to recent ratings can speed up the convergence,
%but assigning too high importance scores
%can slow down the convergence.
 
\begin{figure*}[htb] 
\centering
\subfigure[Recency awareness $c = 0$]{
\centering
\includegraphics[width=0.3\textwidth]{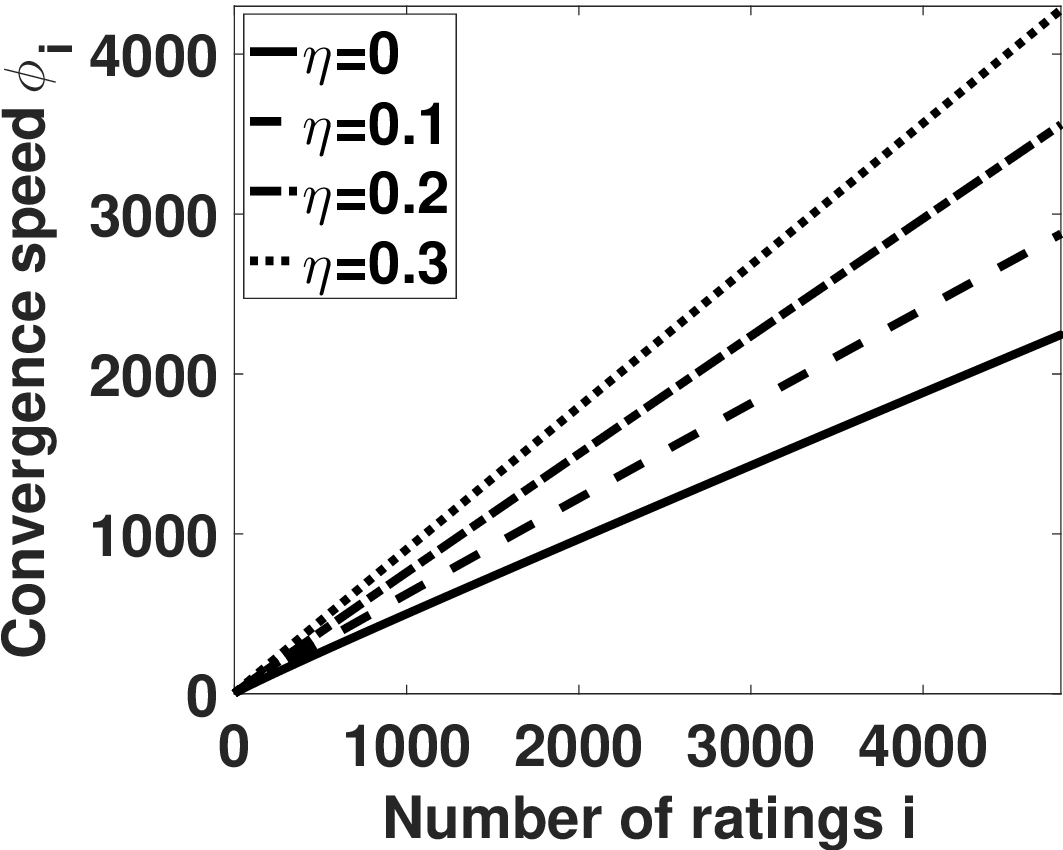}
\label{fig:ImpEtaSpeedConc0}
}
\subfigure[Recency awareness $c = 0.2$]{
\centering
\includegraphics[width=0.3\textwidth]{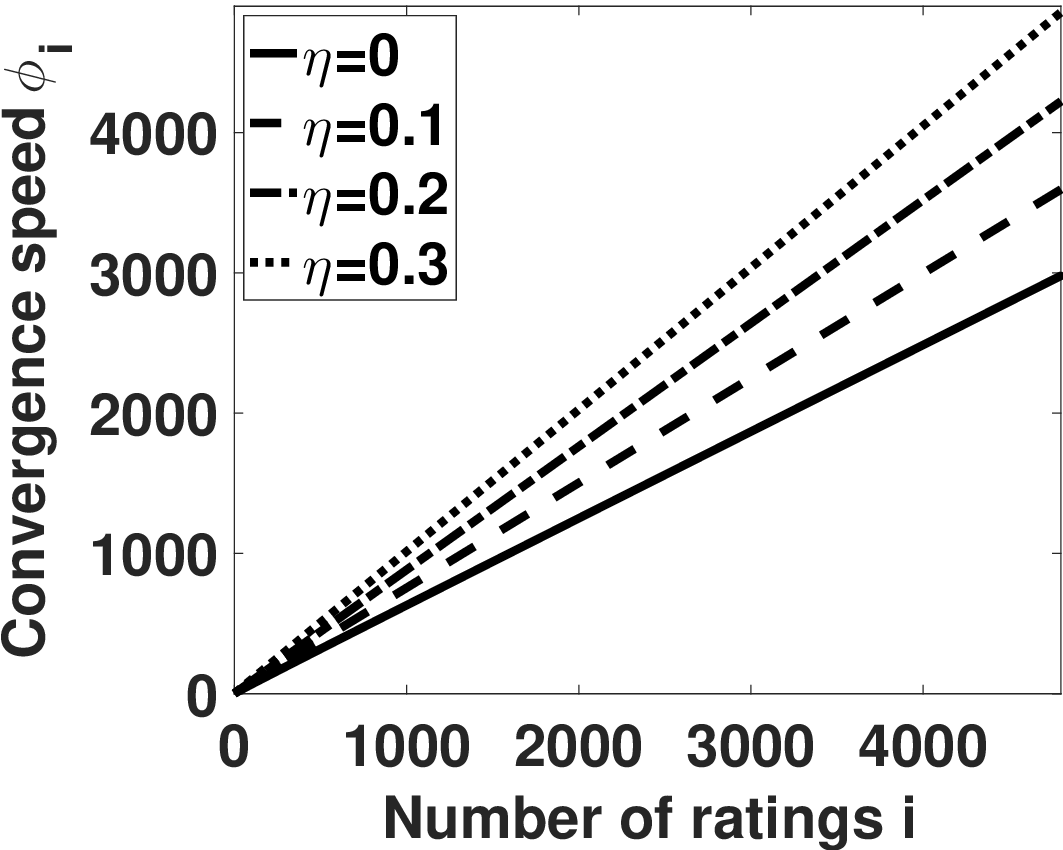}
\label{fig:ImpEtaSpeedConc02}
}
\subfigure[Recency awareness $c = 5$]{
\centering
\includegraphics[width=0.3\textwidth]{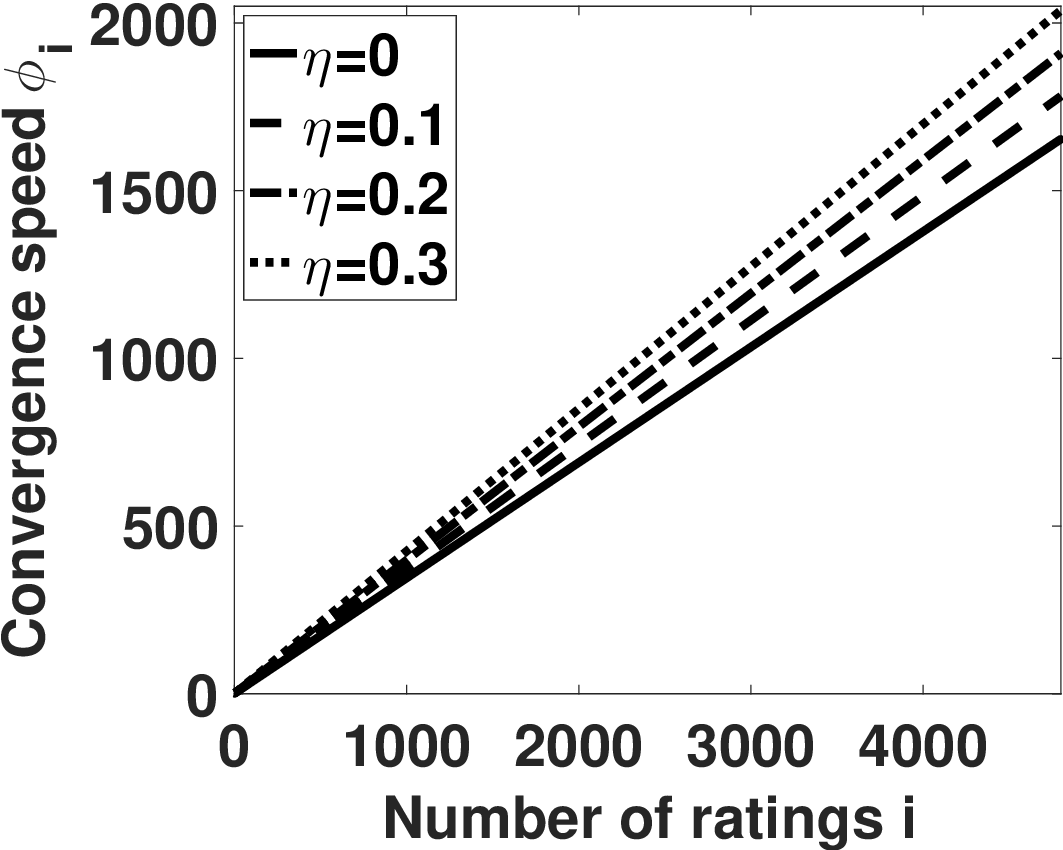}
\label{fig:ImpEtaSpeedConc5}
}
\caption{Impact of review selection mechanisms
(i.e., $\eta$) on the speed of convergence,
where $\gamma \!=\! 0.4$ and $\phi_i$ denotes
$\phi_i (\mathbf{w}, \bm{\eta}, \bm{\gamma})$. }
\label{fig:ImpEtaSpeedCon}
\end{figure*}

\noindent
{\bf Impact of review selection mechanism ($\eta$).}
Recall that each review selection mechanism is modeled
via a parameter $\eta$ capturing the accuracy.
Figure \ref{fig:ImpEtaSpeedCon} shows $\phi_i$ as we vary $\eta$
from 0 to 0.3.
One can observe that $\phi_i$ increases in $\eta$.
This shows that increasing the accuracy of review selection mechanism
can speed up the convergence.
The improvement of $\phi_i$ becomes small,
as we increase $c$.
This implies that an accurate review selection mechanism can improve
the convergence speed significantly
only when the strength of recency awareness
of an aggregation rule is not very strong.
%In such situation, review selection mechanism
%would not be desirable, especially when the associated cost is high.

\begin{figure*}[htb] 
\centering
\subfigure[Recency awareness $c = 0$]{
\centering
\includegraphics[width=0.3\textwidth]{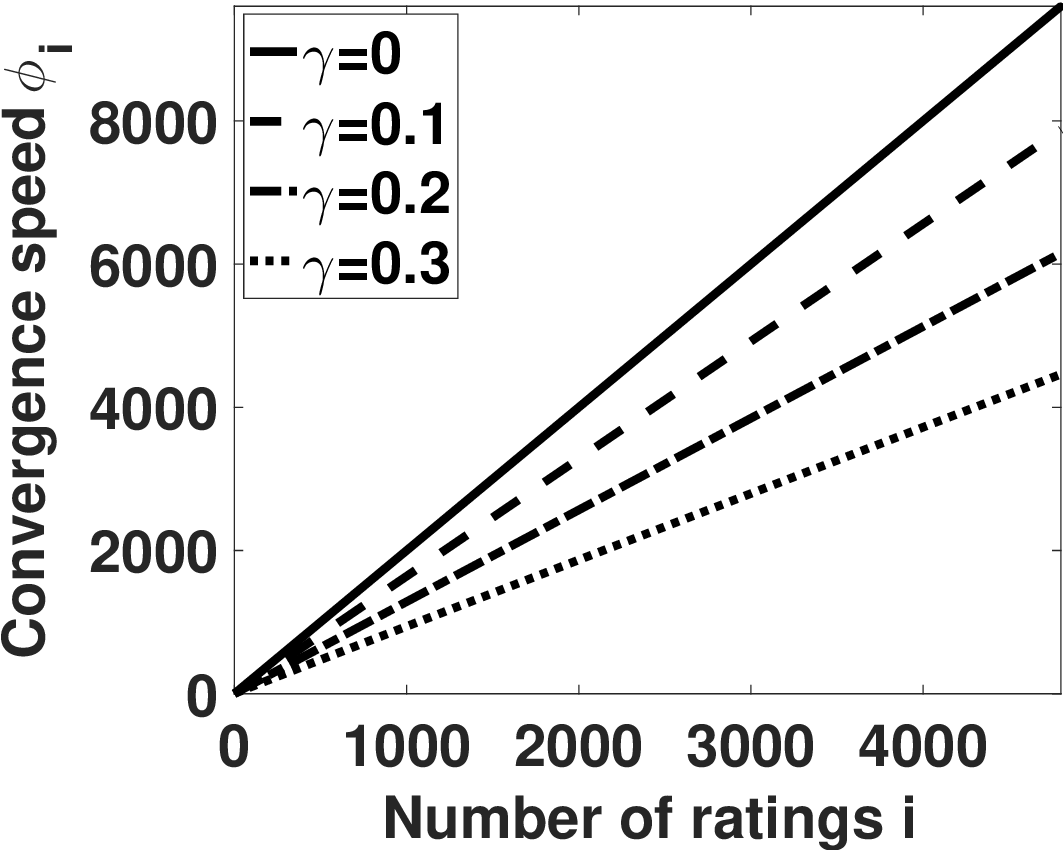}
\label{fig:ImpGamSpeedConc0}
} 
\subfigure[Recency awareness $c = 0.2$]{
\centering
\includegraphics[width=0.3\textwidth]{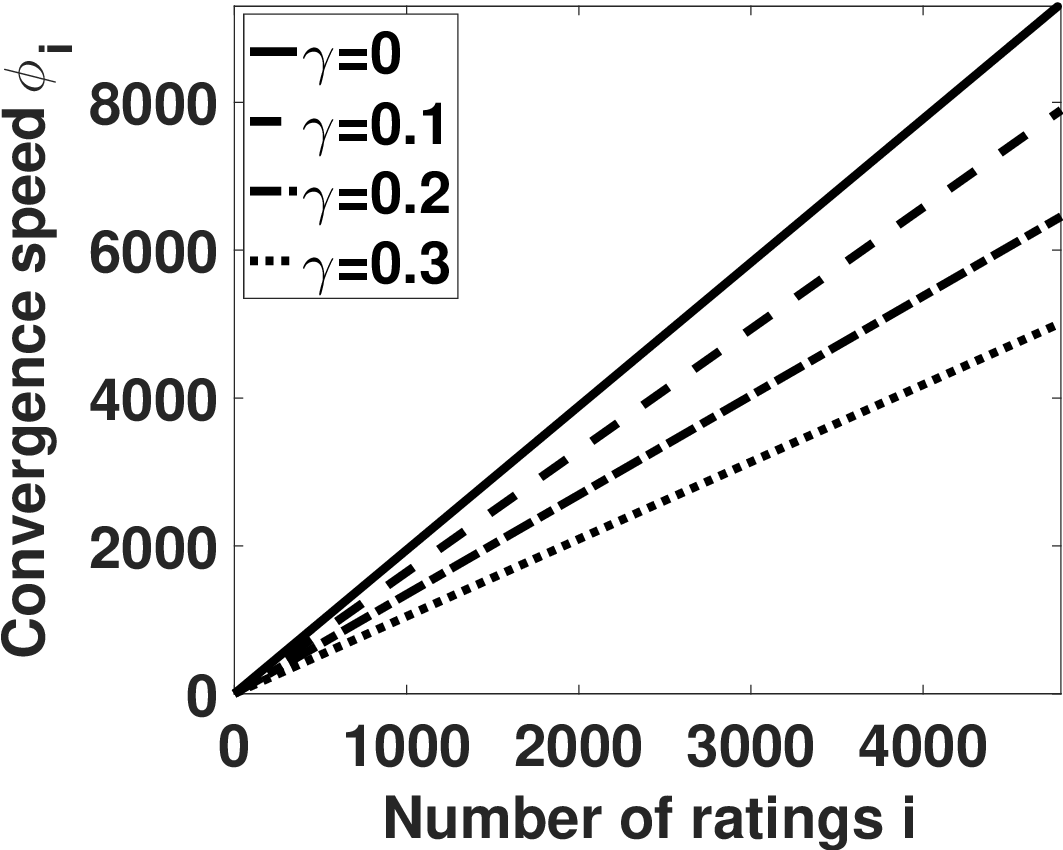}
\label{fig:ImpGamSpeedConc02}
} 
\subfigure[Recency awareness $c = 5$]{
\centering
\includegraphics[width=0.3\textwidth]{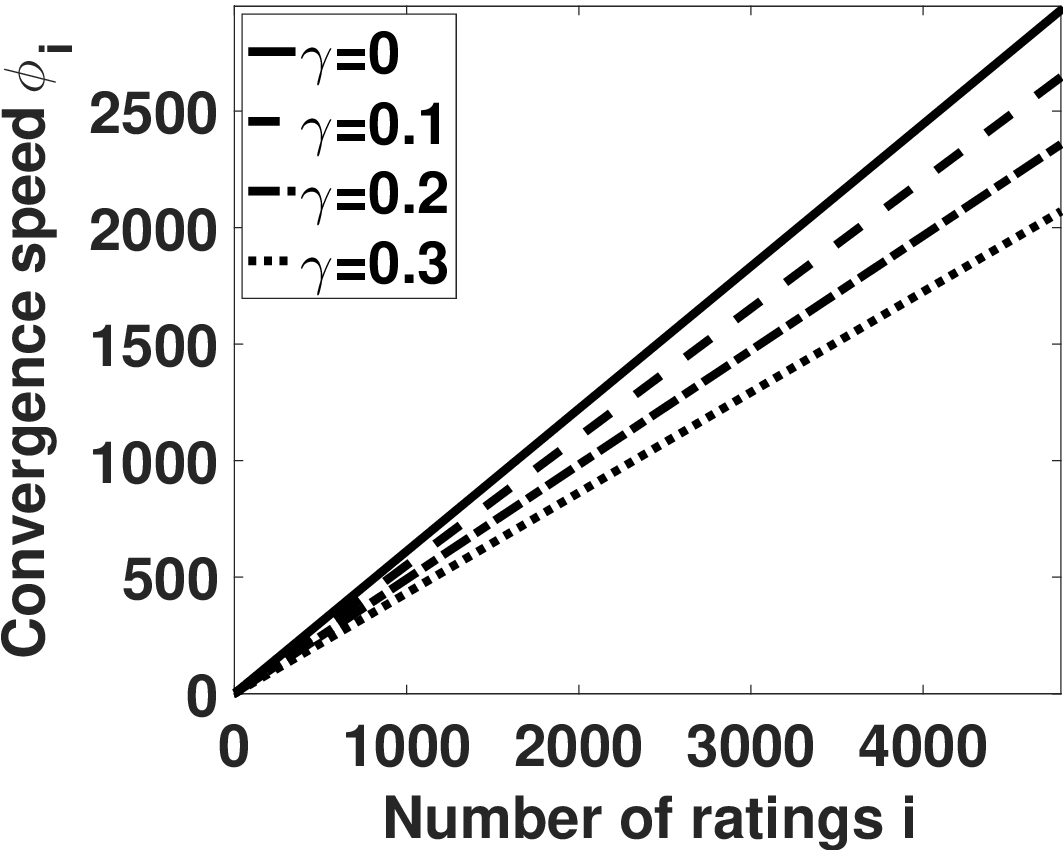}
\label{fig:ImpGamSpeedConc5}
}
\caption{Impact of herding effects (i.e., $\gamma$)
on the speed of convergence, where
$\eta = 0.1$ and $\phi_i$ denotes
$\phi_i (\mathbf{w}, \bm{\eta}, \bm{\gamma})$. }
\label{fig:ImpGamSpeedCon}
\end{figure*}

\noindent
{\bf Impact of herding effects ($\gamma$).}
Recall that the herding effects is modeled via
a parameter $\gamma$ capturing the strength of herding.
Figure \ref{fig:ImpGamSpeedCon} shows $\phi_i$ as we vary $\gamma$
from 0 to 0.3.
One can observe that $\phi_i$ decreases in $\gamma$.
This shows that as users become more prone to herding effects,
the speed of convergence slows down.
The decrease of $\phi_i$ becomes small,
as we increase $c$.
In other words, the rating aggregation rules having higher
strength of recency awareness are more robust
against herding effects.

\noindent
{\bf Lessons learned. }
%The speed of convergence is exponential in the number of ratings.
The recency aware aggregation rule
can speed up the convergence over the simple unweighted average rule,
but the strength of recency awareness should not be too strong.
Increasing the the accuracy of a review selection mechanism
can always improve the convergence speed,
and this improvement decreases as the strength of recency
awareness increases.
Herding effects slow down the speed of convergence,
but the aggregation rule with stronger strength of recency awareness
can be more robust (in terms of convergence speed) against herding effects. 

\subsection{\bf Evaluating the Inference Algorithm}

%Now we evaluate the accuracy of our parameter inference algorithm
%(stated in Section \ref{subsec:LinApprox}).
Recall that $\bm{\alpha}_N$ and $\tilde{\gamma}_N$
denote the inferred model parameters.
We aim to study the accuracy of
 $\bm{\alpha}_N$ and $\tilde{\gamma}_N$
in estimating the best linear approximation
denoted by $\bm{\alpha}^\ast_{lin}$ and $\tilde{\gamma}^\ast_{lin}$.
In general, the best linear approximation
$\bm{\alpha}^\ast_{lin}$ and $\tilde{\gamma}^\ast_{lin}$
is determined by the specific form of
$\bm{\alpha}$ and $\bm{\theta}_i$.
For simplicity, here we consider approximating a linear model,
which is specified in Section \ref{subsec:expSynConSp},
i.e., $\bm{\theta}_i$ derived in Equation (\ref{example:Thetai}),
$\eta_i = \eta$ and $\gamma_i = \gamma$.
Then the best linear approximation for this linear model is
$
\bm{\alpha}^\ast_{lin}
=\bm{\alpha},
\tilde{\gamma}^\ast_{lin}
=(1-\eta) \gamma.
$
We define the relative estimation error as
\begin{align}
\label{eq:ErroMetric}
&
E_\alpha
{\triangleq}
\frac{ 
\mathbb{E}
\left[ \left\|
\bm{\alpha}_N - \bm{\alpha}^\ast_{lin}
\right\|_1
\right]
}{
\left\| \bm{\alpha}^\ast_{lin} \right\|_1
},
&&
E_\gamma
{\triangleq}
\frac{ \mathbb{E}
\left[ \left|
\tilde{\gamma}_N - \tilde{\gamma}^\ast_{lin}
\right| \right]
}{
\tilde{\gamma}^\ast_{lin}
}.
\end{align}
We calculate $E_\alpha$ and $E_\gamma $ via the Monte Carlo simulation.
For each round of simulation,
we use our model
(with parameters $\bm{\alpha}$,
$\eta$, $\gamma$  and $c$,
which will be specified later)
to generate $N$ ratings.
Inputting these $N$ ratings
to our inference algorithm stated in Section \ref{subsec:LinApprox},
we obtain $\bm{\alpha}_N$ and $\tilde{\gamma}_N$
for one round.
Then, we compute one sample of
$E_\alpha$ and $E_\gamma $ via Equation (\ref{eq:ErroMetric}).
We repeat this process for multiple rounds
to obtain multiple samples of $E_\alpha$ and $E_\gamma $.
Lastly, we use the average of these samples
to estimate  $E_\alpha$ and $E_\gamma $.

\begin{figure}[htb]
\centering
\subfigure[Honest rating]{
\centering
\includegraphics[width=0.22\textwidth]{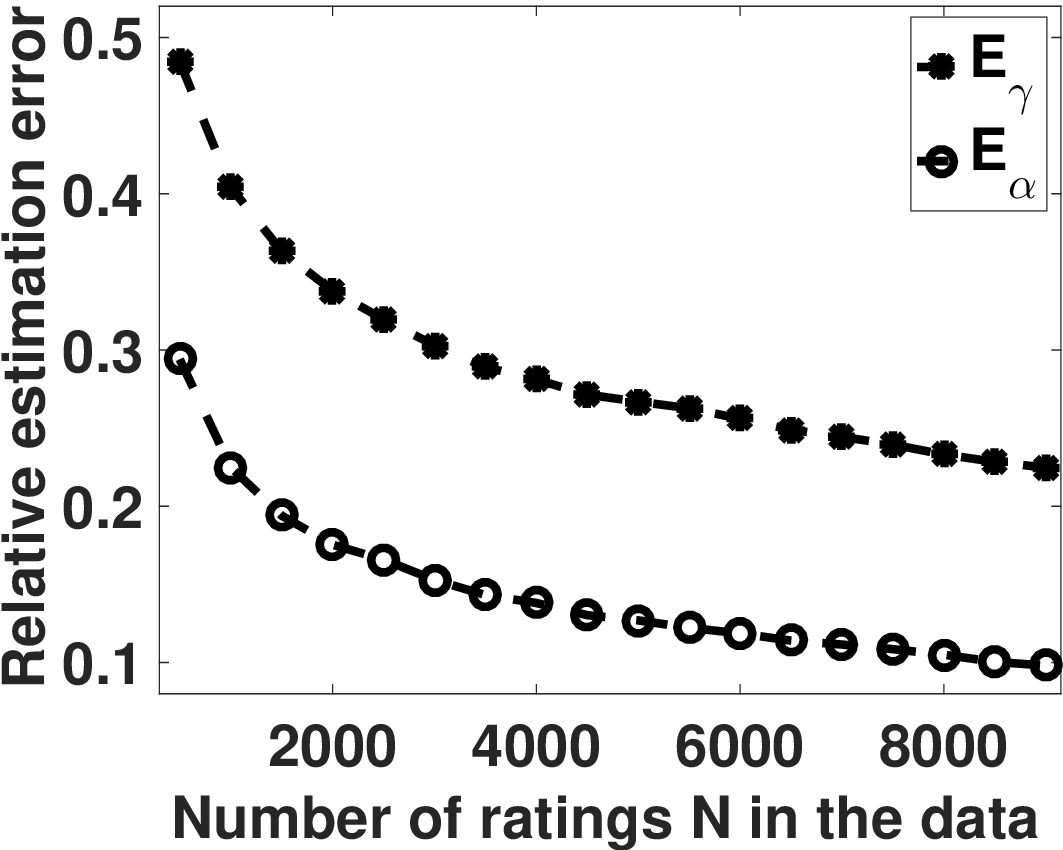}
\label{fig:ExpSynEstAcuInfAlgoHonest}
}
%\hspace{-0.08 in}
\subfigure[Misb. $\mathcal{I}\!=\!\{51,\ldots,100\}$]{
\centering
\includegraphics[width=0.22\textwidth]{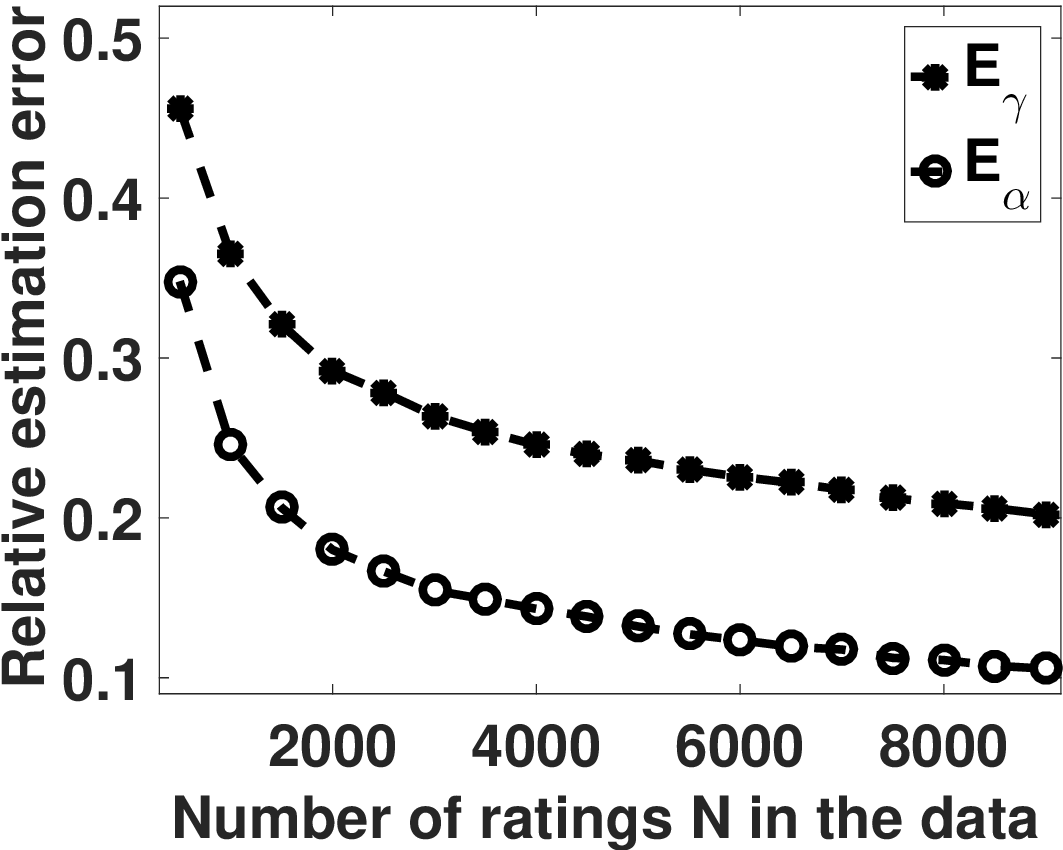}
\label{fig:ExpSynEstAcuInfAlgoMisb}
}
\caption{Estimation error $E_\gamma, E_\alpha$ of our inference algo. }
\label{fig:ExpSynEstAcuInfAlgo}
\end{figure}

Figure \ref{fig:ExpSynEstAcuInfAlgo} shows
$E_\alpha$ and $E_\gamma $ across the number of ratings $N$
under both the honest rating and misbehaving rating scenarios,
where we set
$\bm{\alpha}
=[0, 0,0.1,0.4,0.5]
$,
$(1-\eta) \gamma = 0.8$  and
$c=0.$
We consider a large effective strength of herding effects
$(1-\eta) \gamma=0.8$,
because the robustness of our inference algorithm
(against misbehaving ratings) under small strength
can be implied by this large strength.
Figure \ref{fig:ExpSynEstAcuInfAlgoHonest}
shows that under the honest rating scenario,
both $E_\alpha$ and $E_\gamma $ decrease in
the number of rating $N$.
This implies that the accuracy of our inference algorithm
increases as more ratings are observed.
Furthermore, the curve of $E_\gamma $ lies above
$E_\alpha$.
This means that our inference algorithm has a higher
accuracy in estimating the ground-truth collective opinion
$\bm{\alpha}^\ast_{lin}$
than the effective strength of herding effects $\tilde{\gamma}^\ast_{lin}$.
The error $E_\alpha$ and $E_\gamma$ can be as
small as 10\% and 20\% respectively
with thousands of ratings.
Figure \ref{fig:ExpSynEstAcuInfAlgoMisb} presents
$E_\alpha$ and $E_\gamma$ when we inject
50 misbehaving ratings (i.e., $\mathcal{I}=\{51,\ldots,100\}$) of 5.
Compared to Figure \ref{fig:ExpSynEstAcuInfAlgoHonest},
one can observe that our inference algorithm has similar accuracy
as that of the honest rating case.
This implies that our inference algorithm is robust against
misbehaving ratings.

\noindent
{\bf Lessons learned. }
Our inference algorithm can achieve a high accuracy
under thousands of ratings and it is robust against misbehaving
ratings.

\section{\bf Experiments on Real Data}
\label{sec:ExpRealData}

We conduct experiments on real-world
online ratings from Amazon and TripAdvisor.
We identify recency aware rating aggregation rules,
which  improve the convergence speed in Amazon and TripAdvisor
by 41\% and 62\% respectively.

%In this section we conduct experiments on synthetic data to
%study the impact of $\mathbf{w}, \bm{\eta}$ and $\bm{\gamma}$
%on the speed of convergence
%$\phi_i (\mathbf{w}, \bm{\eta}, \bm{\gamma})$.
%Through this we uncover fundamental understandings to
%guide online rating system operators to shepherd online product ratings.

{%\color{blue}
%\subsection{Experiment Settings} 
\subsection{\bf Datasets and Parameter Inference.}
 
We use the ratings of 32,888 products in Amazon and
11,543 hotels in TripAdvisor crawled in 2013.
From the rating dataset we select the items
(i.e., products in Amazon or hotels in TripAdvisor)
with at least 2,000 ratings to attain a balance between
inference accuracy and dataset scale.
In total, 284 products in the Amazon dataset and
111 hotels in TripAdvisor dataset are selected.
Note that Amazon and TripAdvisor use
the simple unweighted average rule, i.e., $w_i = i^c, c=0$.
We apply the Algorithm stated in Section \ref{sec:Inference}
to infer $\bm{\alpha}_N$ and $\tilde{\gamma}_N$
for each selected item.   

\begin{figure}[htb]
\centering 
\includegraphics[width=0.36\textwidth]{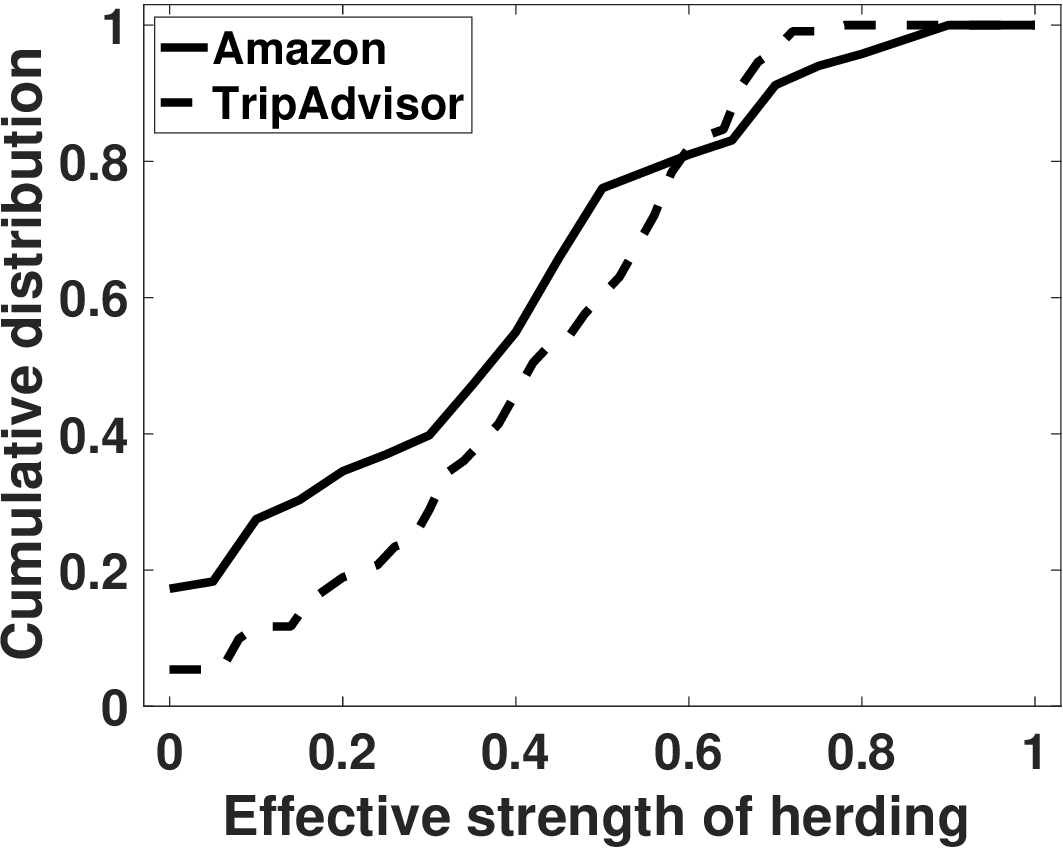}
\caption{Cumulative distribution of $\tilde{\gamma}_N$ across items.}
\label{fig:ExpRealInfParameter}
\end{figure}

Figure \ref{fig:ExpRealInfParameter} shows the cumulative distribution
(CDF) of the inferred effective herding strength $\tilde{\gamma}_N$
across items.
One can observe that the cumulative distribution curve
of Amazon lies above that of TripAdvisor roughly.   
The average of the inferred effective herding strengths
for Amazon and TripAdvisor are
$0.366$ and $0.416$ respectively.
This implies that users in Tripadvisor are 
\textit{more likely to follow the crowd} 
in providing ratings.  
 
\subsection{\bf Rating Aggregation Rules and Implications }
 
We consider $w_i = i^c$.
Figure \ref{fig:ExpReal} shows the value of $\phi_i$
as we vary $c$ from 0 to 4.
Note that the Amazon and TripAdvisor practice $c=0$,
i.e., unweighted aggregation rule.
One can observe that we can improve the speed of convergence in Amzaon
by
$
3919/2785 -1
= 41\%
$
using a recency aware aggregation rule with $c=0.5$
and speed up the convergence in TripAdvisor by
$
3266/2011 - 1
=62\%
$
using a recency aware aggregation rule with $c=0.8$.
In other words, using these aggregation rules,
we can \textit{reveal the ground-truth quality of products with the same
accuracy by using significantly less ratings}
in both Amazon and TripAdvisor.

\begin{figure}[htb]
%\vspace{-0.06 in}
\centering
\subfigure[Amazon (overall $\tilde{\gamma} \!=\! 0.366$)]{
\centering
\includegraphics[width=0.22\textwidth]{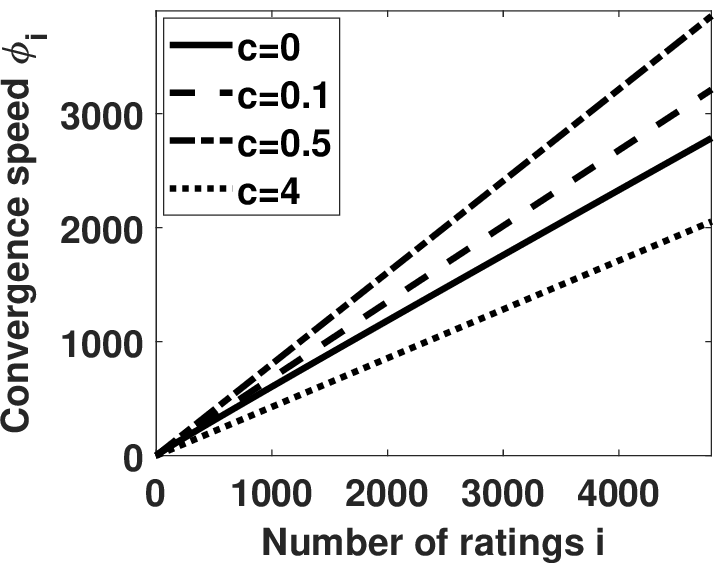}
\label{fig:ExpRealAmazon}
}
%\hspace{-0.08 in}
\subfigure[TripAdv. (overall $\tilde{\gamma} \!=\! 0.416$)]{
\centering
\includegraphics[width=0.22\textwidth]{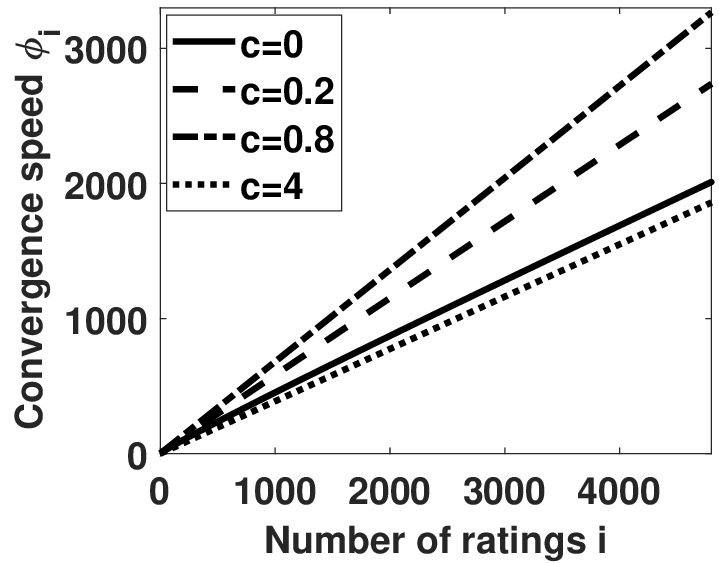}
\label{fig:ExpRealTripAdvisor}
}
%\vspace{-0.06 in}
\caption{Speed of convergence in Amazon and Tripadvisor. }
\label{fig:ExpReal}
\end{figure}

\noindent
{\bf Lessons learned. }
We identify appropriate recency aware aggregation rules,
which can improve the speed of convergence in Amazon and TripAdvisor
by 41\% and 62\%.
}\section{\bf Experiments on Real Data}
\label{sec:ExpRealData}

We conduct experiments on real-world
online ratings from Amazon and TripAdvisor.
We identify recency aware rating aggregation rules,
which  improve the convergence speed in Amazon and TripAdvisor
by 41\% and 62\% respectively.

%In this section we conduct experiments on synthetic data to
%study the impact of $\mathbf{w}, \bm{\eta}$ and $\bm{\gamma}$
%on the speed of convergence
%$\phi_i (\mathbf{w}, \bm{\eta}, \bm{\gamma})$.
%Through this we uncover fundamental understandings to
%guide online rating system operators to shepherd online product ratings.

{%\color{blue}
%\subsection{Experiment Settings} 
\subsection{\bf Datasets and Parameter Inference.}
 
We use the ratings of 32,888 products in Amazon and
11,543 hotels in TripAdvisor crawled in 2013.
From the rating dataset we select the items
(i.e., products in Amazon or hotels in TripAdvisor)
with at least 2,000 ratings to attain a balance between
inference accuracy and dataset scale.
In total, 284 products in the Amazon dataset and
111 hotels in TripAdvisor dataset are selected.
Note that Amazon and TripAdvisor use
the simple unweighted average rule, i.e., $w_i = i^c, c=0$.
We apply the Algorithm stated in Section \ref{sec:Inference}
to infer $\bm{\alpha}_N$ and $\tilde{\gamma}_N$
for each selected item.   

\begin{figure}[htb]
\centering 
\includegraphics[width=0.36\textwidth]{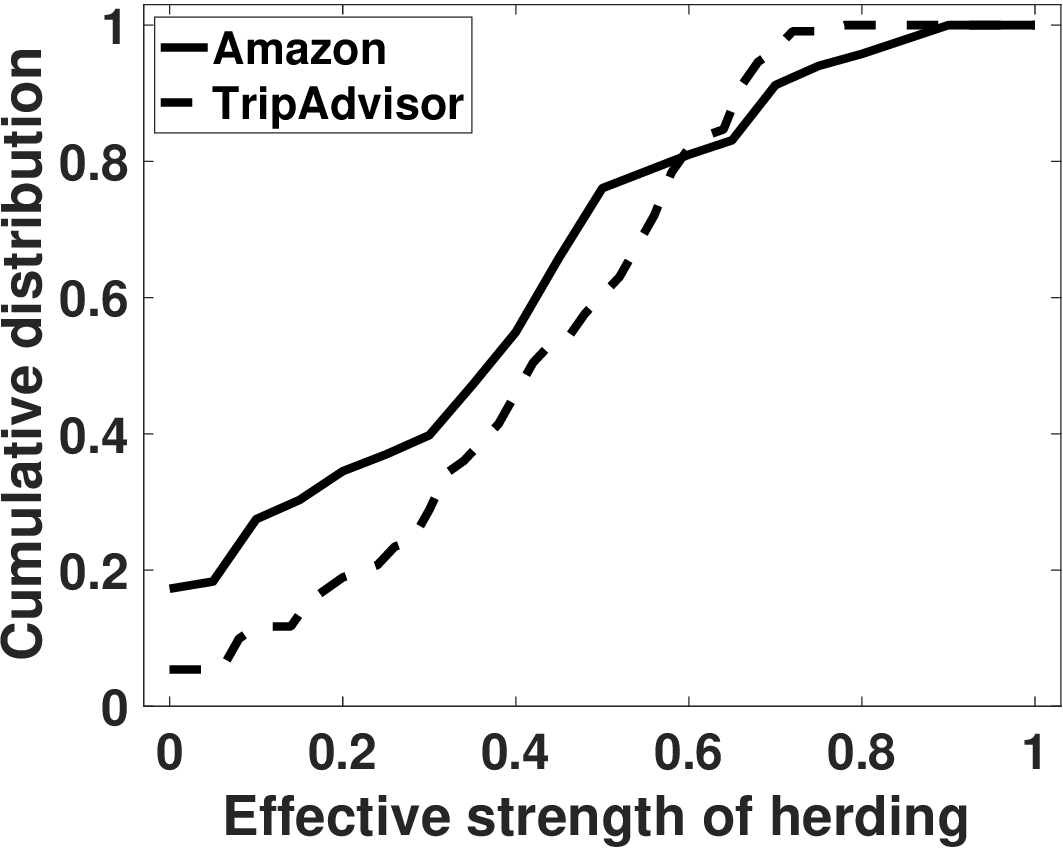}
\caption{Cumulative distribution of $\tilde{\gamma}_N$ across items.}
\label{fig:ExpRealInfParameter}
\end{figure}

Figure \ref{fig:ExpRealInfParameter} shows the cumulative distribution
(CDF) of the inferred effective herding strength $\tilde{\gamma}_N$
across items.
One can observe that the cumulative distribution curve
of Amazon lies above that of TripAdvisor roughly.   
The average of the inferred effective herding strengths
for Amazon and TripAdvisor are
$0.366$ and $0.416$ respectively.
This implies that users in Tripadvisor are 
\textit{more likely to follow the crowd} 
in providing ratings.  
 
\subsection{\bf Rating Aggregation Rules and Implications }
 
We consider $w_i = i^c$.
Figure \ref{fig:ExpReal} shows the value of $\phi_i$
as we vary $c$ from 0 to 4.
Note that the Amazon and TripAdvisor practice $c=0$,
i.e., unweighted aggregation rule.
One can observe that we can improve the speed of convergence in Amzaon
by
$
3919/2785 -1
= 41\%
$
using a recency aware aggregation rule with $c=0.5$
and speed up the convergence in TripAdvisor by
$
3266/2011 - 1
=62\%
$
using a recency aware aggregation rule with $c=0.8$.
In other words, using these aggregation rules,
we can \textit{reveal the ground-truth quality of products with the same
accuracy by using significantly less ratings}
in both Amazon and TripAdvisor.

\begin{figure}[htb]
%\vspace{-0.06 in}
\centering
\subfigure[Amazon (overall $\tilde{\gamma} \!=\! 0.366$)]{
\centering
\includegraphics[width=0.22\textwidth]{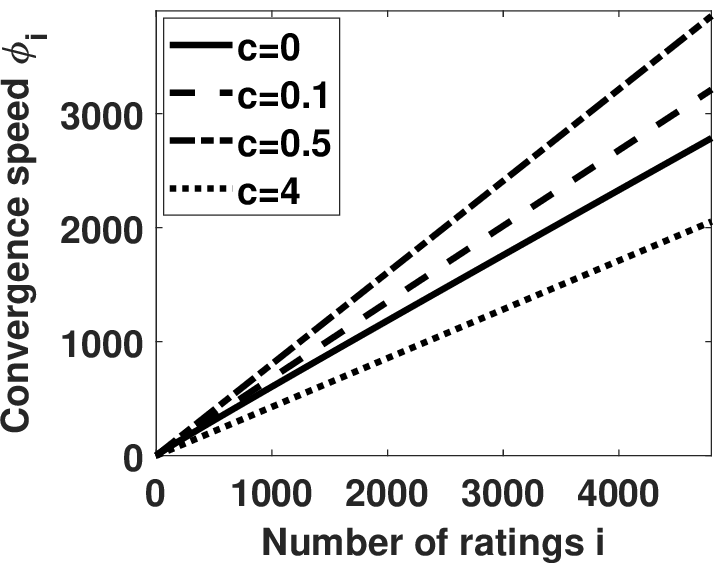}
\label{fig:ExpRealAmazon}
}
%\hspace{-0.08 in}
\subfigure[TripAdv. (overall $\tilde{\gamma} \!=\! 0.416$)]{
\centering
\includegraphics[width=0.22\textwidth]{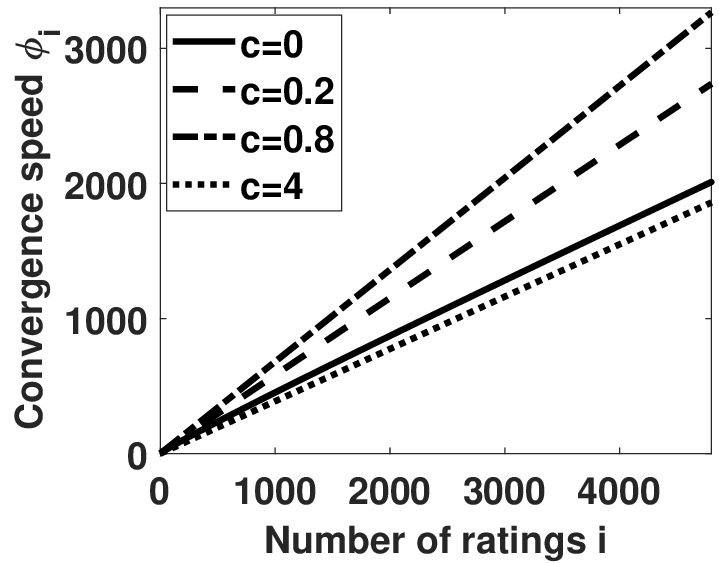}
\label{fig:ExpRealTripAdvisor}
}
%\vspace{-0.06 in}
\caption{Speed of convergence in Amazon and Tripadvisor. }
\label{fig:ExpReal}
\end{figure}

\noindent
{\bf Lessons learned. }
We identify appropriate recency aware aggregation rules,
which can improve the speed of convergence in Amazon and TripAdvisor
by 41\% and 62\%.
}

\section{\bf Application: Rating Prediction}
\label{sec:ratingPrediction}

To demonstrate the versatility of our model, 
we parametrize our model to predict subsequent ratings.  
We apply regularized least square to infer model parameters.  
Extensive experiments on four datasets demonstrate 
that our model can improve the accuracy of 
Herd \cite{} and the HIALF \cite{}.

\subsection{\bf Applying Our Models to Rating Prediction}

We consider the following rating prediction problem: 
\textit{
given a set of historical ratings of items, 
predict subsequent ratings of items.  
} 
To apply our model to address this problem, 
we next first parameterize our model 
and then infer model parameters.  

\noindent
Let $r_{u,i} \in \mathcal{M}$ denote the ubiased rating of user $u \in \mathcal{U}$ 
toward $i \in \mathcal{I}$. Namely, $r_{u,i}$ characterizes user $u$'s intrinsic overall
opinion toward item $i$

\noindent
{\bf Modeling initial opinion formation. }
Now we model the initial opinion formation, 
i.e., the function $\Gamma_u ( \mathcal{H}_{i,k} )$.   
Users may form initial opinions from the aggregation of historical ratings or 
a small number of latest ratings.  
We first consider the case that users form initial opinions from 
the aggregation of historical ratings.  
Denote the collective opinion summarized 
from the rating history $\mathcal{H}_{i,k}$ as 
\[
\bm{h}_{i,k}
\!\triangleq\!
[h_{i, k, 1}, \ldots, h_{i, k, M}], 
\]
where $h_{i,k,m} \in [0,1]$ and 
$\sum_{m \in \mathcal{M}} h_{i,k,m} =1$.   
The $\bm{h}_{i,k}$ is public to all users.  
We consider a \textit{class} of weighted aggregation rules
to summarize historical ratings: 
\begin{align} \label{eq:beta}
&
h_{i, k, m}
=
\frac{\sum^k_{j = 1} \alpha_j \mathbb{I}_{\{R_{i,j} = m\}}}
{\sum^k_{j=1} \alpha_j},
\end{align}
where $\alpha_j \in \mathbb{R}_+$ denotes 
the weight associated with $j$-th rating,
and $\mathbb{I}$ is an indicator function.  
For example, $\alpha_j \!=\! 1, \forall j$, is deployed in Amazon and TripAdvisor, 
which corresponds to ``\textit{average rating rule}''.  
Under this average rating rule, we have 
$
h_{i, k, m}
\!=\!
\sum^k_{j=1} \mathbb{I}_{\{R_{i,j} = m\}} / k
$,
which is the fraction of historical ratings equal $m$.  
Note that $\bm{h}_{i,k}$ is displayed to all users.  
We capture the aggregate opinion heterogeneity 
in initial opinion formation as follows:  
\begin{align}
\Gamma_u ( \mathcal{H}_{i,k} )
= 
\frac{
\sum_{ m \in \mathcal{M} } m \beta_m h_{i,k,m} 
}{
\sum_{ m \in \mathcal{M} } \beta_m h_{i,k,m}
},
\label{eq:Model:IniOpOrder}
\end{align}
where the weight $\beta_m \in [0,1]$ models how a user weighs 
the opinion associated with each rating level.  
For example, a user may assign a large weight to low ratings representing 
that she is sensitive to negative opinions.  
There are several possible parametric forms of the weights: 
\[
\beta_m 
= 
\exp ( - \gamma m ), 
\,\,\,
\beta_m 
= 
m^\gamma, 
\,\,\,
\beta_m 
= 
\ln ( m^\gamma + 1).
\]
where $\gamma \in \mathbb{R}$.  

Now, we consider the case that users form initial opinions from 
a small number of latest ratings.  
Let $n \in \mathbb{N}_+$ denote the number of latest ratings 
that users refer to for initial opinion formation.  
We capture rating recency in initial opinion formation as: 
\begin{align}
\Gamma_u ( \mathcal{H}_{i,k} )
= 
\frac{
\sum_{j =1}^{n} 
\eta_{k,j}  
R_{i,k-j+1} 
\mathbb{I}_{ \{ k-j+1 \geq 1 \}}
}{
\sum_{j =1}^{n}  
\eta_{k,j}  
\mathbb{I}_{ \{ k-j+1 \geq 1 \}}
},
\label{eq:Model:IniOpinRecency}
\end{align}
where the weight satisfies $\eta_{k,j} \in \mathbb{R}_+$ 
and we set $R_{i,j} = 0$ for all $j <1$ by default.   
There are several possible forms of the weight $\eta_{k,j}$.  
One can use the following forms of $\eta_{k,j}$ to model 
arrival order aware initial opinion formation: 
\[
\eta_{k,j} 
= 
\exp ( - \zeta j ), 
\,\,\,
\eta_{k,j} 
= 
\frac{1}{ j^{ \zeta } }, 
\,\,\,
\eta_{k,j} 
= 
\frac{1}{ \zeta \ln( j + 1 ) }.  
\]
where $\zeta \in \mathbb{R}$.   
One can use the following forms of $\eta_{k,j}$ to model 
arrival time stamp aware initial opinion formation:  
\begin{align*}
& 
\eta_{k,j} 
= 
\exp 
\left( 
- \rho ( t_u - t_{i,k+1-j} )^{ \delta } 
\right), 
\,\,\,
\eta_{k,j}  
= 
\frac{1}{ ( t_u - t_{i,k+1-j} )^{ \rho }  }
\\
& 
\eta_{k,j}  
= 
\frac{1}{ \rho \ln ( t_u - t_{i,k+1-j} + 1)   }. 
\end{align*}
where $t_u$ denotes the arrival time of user $u$, 
$\delta \in \mathbb{R}, \rho \in \mathbb{R}$, 
we set $t_{i,j} = 0$ by default for all $j < 1$.  
  
\noindent
{\bf Model parameterization.}  
The Herd algorithm was proposed by Wang \textit{et al.} \cite{}, 
which is the first algorithm exploring Herd effects to improve rating prediction.  
We will use Herd as a major comparison baseline.  
For fair comparison with Herd \cite{}, 
we parameterize $r_{u,i}$ via the classical latent factor (LF) model \cite{}: 
\begin{equation}
r_{u,i} 
= 
\bm{x}_u^T \bm{y}_i + b_i + b_u + g. 
\label{eq:RatPred:ParametrizeRat}
\end{equation}  
In Equation (\ref{eq:RatPred:ParametrizeRat}), 
$\bm{x}_u \in \mathbb{R}^{\kappa}$ and $\bm{y}_i \in \mathbb{R}^{\kappa}$ 
represent vectors of latent features for user $u$ and product $p$, 
where $\kappa \in \mathbb{N}_+$.  
The $b_u \in \mathbb{R}$ and $b_i \in \mathbb{R}$ model user and product bias respectively.   
The $g \in \mathbb{R}$ models the constant shift or residual.   
All parameters in Equation (\ref{eq:RatPred:ParametrizeRat}) 
are unknown and will be inferred from the data.  
Similarily, we parametrize the strength of herding effects as:
\[
s_{u, k} 
=
\frac{2}{ 1 + \exp ( - {s_{\mathcal{U}}}^2 \times k)} - 1
\]
where $s_{\mathcal{U}} \in \mathbb{R}$ is unknown parameters 
to be inferred from data.  
Note this parametrization method was used in \cite{}, 
and we choose it for fair comparison with Herd.   
We will show that our model can outperform all the baselines.  
This implies that if one selects them with finer tuning, 
our model can achieve better performance.  
We summarize all parameters to be inferred from data as 
\[
\mathcal{Z} 
\triangleq
\left\{ 
g,\{{b_i},{\boldsymbol{y}_i}\}_{i\in\mathcal{I}}, \{{b_u}, {\boldsymbol{x}_u}\}_{u\in\mathcal{U}},s_{\mathcal{U}}
\right\}
\]
All the hyper parameters will be given before model training.  

\noindent
{\bf Model inference.}  
Consider a training rating dataset, 
in which product $p$ has $K_i \geq 0$ historical ratings.  
We aim to infer $\mathcal{Z} $ from the training rating dataset.  
In particular, we use regularized least square method to infer $\mathcal{Z} $:    
\begin{align*} 
\min_{ \mathcal{Z} }
\sum\nolimits_{ i \in \mathcal{I} }  
\sum\nolimits^{K_i}_{k=1}  
&
\bigg[
r_{u,i,k}\\
&
-\bigg(
(\frac{2}{ 1 + \exp ( - s_{\mathcal{U}} \times k)} - 1) \times \Gamma_u ( \mathcal{H}_{i,k} )\\
&
+ (1 -(\frac{2}{ 1 + \exp ( - s_{\mathcal{U}} \times k)} - 1) ) \\
&
\times (\bm{x}_u^T \bm{y}_i + b_i + b_u + g) 
\bigg)
\bigg]^2\\
&
+ \lambda_{rec}(b_u^2 + b_i^2 + ||\bm{x}_u||_2^2 + ||\bm{y}_i||_2^2)\\
&
+ \lambda_s(s_{\mathcal{U}}^2)
\end{align*} 
We use stochastic gradient descent (SGD) algorithm 
to learn model parameters $\mathcal{Z} $.   
Note that SGD was widely used in previous works 
to improve training efficiency \cite{}

\noindent
{\bf Rating prediction. }
Let $\widehat{\mathcal{Z} }$ denote the inferred model parameter set.  
To illustrate, suppose we are going to predict the $i$-th rating of item $i$, 
i.e., $R_{i,k}$.  
Note that $u_{k,i}$ is the user who assign the rating $R_{i,k}$.  
We are given the ID of the user denoted by $u_{k,i}$.  
Then we predict the rating $R_{i,k}$ as $\widehat{r}_{u_{k,i},i}$, 
which are computed using our model with the inferred 
parameters $\widehat{\mathcal{Z} }$.   
We denote our rating prediction method as R-HE.

\subsection{\bf Evaluation Settings}

\noindent
{\bf The dataset. }
We use four public datasets to evaluate the accuracy of our AC-RP method, 
whose overall statistics are summarized in Table \ref{tb:RatPred:rawdata}.  
The dataset from Amazon was published in \cite{} and it 
contains historical ratings of movies in Amazon.  
The dataset from Google Local was published in \cite{} 
and it contains reviews about businesses from Google Local (Google Maps).  
The dataset from TripAdvisor was published in \cite{}, 
and it contains historical ratings of hotels in TripAdvisor.  
The dataset from Yelp was downloaded from the link \footnote{https://www.yelp.com/dataset}
and it contains historical ratings for restaurants in Yelp.  

\begin{table}[htb]
 \caption{\bf Overall statistics of four datasets}
\begin{tabular}{p{60 pt}||p{50 pt}p{50 pt}p{50 pt}}
\hline
\textbf{category} & \textbf{\# products} & \textbf{\# users} & \textbf{\# ratings} \\ \hline
%Amazon            & 1,465                & 429,910           & 1,035,397           \\
Amazon-movie      & 208,321              & 2,088,620         & 4,607,047           \\
Googlelocal       & 4,567,431            & 3,116,785         & 10,601,852          \\
TripAdvisor       & 1,705                & 623,567           & 871,689             \\
Yelp              & 60,785               & 366,715           & 1,569,264           \\ \hline
\end{tabular}
\label{tb:RatPred:rawdata}
\end{table}

\noindent  
For fair comparison with Herd, 
we use the same method as HIALF \cite{Zhang2019} to extract training dataset and testing dataset.  
Similar with HIALF \cite{}, 
for each item in Table \ref{tb:RatPred:rawdata}, 
we only select items with an medium positive average rating, 
i.e., average rating in the range $[3.9,4.1]$ out.  
For each selected item we extract all its ratings 
and the associated users out.  
We then remove items with less than 50 training ratings, 
to avoid over-fitting.  
Table \ref{tb:RatPreSelectData} summarizes overall statistics of the selected data.  
One can observe that after this selection, 
some datasets still contain around two hundred thousands of users.  
Comparing the number of ratings with the number of  users, 
one can observe that the rating matrix is very sparse.  
Similar with HIALF \cite{}, 
we further aggregate users with twenty ratings (or fifty)
as a big user for Googlelocal and TripAdvisor ( or Amazon-movie and Yelp).  
Furthermore, for each selected item, 
we use its last 25 ratings as test ratings 
and use all other ratings as training ratings.  
We train the models on the training dataset, 
and validate the model on the testing dataset.  

\begin{table}[htb]
 \caption{\bf Overall statistics of selected rating dataset. }
\begin{tabular}{p{60 pt}||p{35 pt}p{35 pt}p{35 pt}}
\hline
\textbf{category} & \textbf{\# products} & \textbf{\# users} & \textbf{\# ratings}  \\ \hline
%Amazon       & 194   & 11,983 & 69,158    \\
Amazon-movie & 1,118 & 198,209 & 271,295   \\
Googlelocal  & 445   & 38,553  & 53,740    \\
TripAdvisor  & 309   & 134,484 & 150,201   \\
Yelp         & 736   & 82,509  & 157,463   \\ \hline
\end{tabular}
\label{tb:RatPreSelectData}
\end{table} 

\noindent
{\bf Comparison baseline \& metrics. } 
We use the root mean squared error (RMSE) to quantify the testing accuracy: 
\begin{equation}
RMSE 
= 
\sqrt{
\frac{
\sum_{ p \in \mathcal{P} }  
\sum^{K_p}_{k=K_p - 24}   
( \widehat{r}_{u_{p,i}, p,i} - R_{p,i} )^2}{25 | \mathcal{P}| }
},
\end{equation}
where $\widehat{r}_{u_{p,i}, p,i}$ denotes an estimation of $r_{u_{p,i}, p,i}$, 
which is computed using our model with the inferred parameters $\widehat{\mathcal{Z}}$.  
We compare our AC-RP method with the following three baselines.  
\begin{itemize}
\item 
HIALF \cite{}.  
The HIALF algorithm was proposed by Zhang \textit{et al.} \cite{}, 
which is the first algorithm exploring assimilate-contrast effects to improve rating prediction.   

\item 
Herd \cite{}. 
The Herd algorithm was proposed by Zhang \textit{et al.}, 
which explores herding effects to improve rating prediction.   
\end{itemize}

\noindent
{\bf Parameter setting.} 
To prevent over fitting, we set appropriate regularization hyper parameters for our models.  
Following similar principle in HIALF \cite{}, 
we consider the regularization hyper parameters summarized in Table \ref{tb:RatPrid:hyper}.   
Following previous work \cite{}, 
we choose the dimension of latent features as $\kappa=5$.  
For the other hyper parameters, we will select them systematically 
and present them with the experiment results.  

\begin{table}[htb]
\caption{\bf Hyper parameter}
\begin{tabular}{ccc}
\toprule
$\lambda_{rec}$&
$\lambda_s$&
$\kappa$\\ \midrule
0.1&
0.01&
5\\ \bottomrule
\end{tabular}
\label{tb:RatPrid:hyper}
\end{table}

\subsection{\bf Rating Recency for Rating Prediction}  

In this section, we evaluate the benefit of exploiting rating recency 
for rating prediction tasks.   
Consider that under rating recency, users form initial 
opinion from $n$ latest ratings.  
We consider the initial opinion formation model derived in 
Equation %(\ref{eq:Model:IniOpinRecency}).  

\noindent
{\bf Impact of $n$}.  
The initial opinion is the simple average of $n$ latest ratings, 
i.e., $\eta_{k,j} = 1, \forall j$.  
Table \ref{tb:RatPred:Impn} shows the RMSE of Herd, HIALF and our B-HE method.   
In Table \ref{tb:RatPred:Impn}, the column $n=20,40,80$ 
corresponds to the RMSE of our B-HE method with $n=20,40,80$.    
One can observe that when $n=20$, 
our method has a smaller RMSE than Herd.   
This statement also holds when $n=40,80$.  
Namely, under some simple selections of $n$, our method has a higher 
rating prediction accuracy than Herd.  
Note that this improvement of rating prediction accuracy 
by exploiting rating recency is supported by survey studies, 
which identified that users tend to read a small number of latest reviews or ratings 
to form initial opinions \cite{}.   
The RMSE of our method varies as we increase $n$ from 20 to 80, 
This implies that $n$ is an important factor for the rating prediction accuracy 
and one needs to exploit the rating recency carefully for rating prediction.  
The above improvement on the rating prediction accuracy is achieved 
at simple selections of $n$.  
One may further improve the rating prediction accuracy by finer tuning of $n$.   
In this experiment, the initial opinion is the simple average of $n$ latest ratings.       
Users may have different weights to different ratings, 
i.e., more weights on more recent ratings.  
In the following, we explore this direction.

\begin{table}[htb]
 \caption{\bf Impact of $n$ on rating prediction ($\eta_{k,j} {=}1, \forall j$). }
\begin{tabular}{p{60 pt}||p{40 pt}p{40 pt}p{40 pt}}
\toprule
\textbf{category} & \textbf{Herd} &  \textbf{HIALF} & \textbf{RobustAVG} \\ \midrule
%Amazon       & 1.4471 & 1.3137 & 1.3134   \\
Amazon-movie & 1.4871  & 1.2209  & 1.1848  \\
Googlelocal  & 1.1412  & 1.0616  & 1.1013  \\
TripAdvisor  & 1.1838 & 0.9208  & 1.1307  \\
Yelp         & 1.4200  & 1.2006  & 1.0059  \\ \bottomrule
\textbf{category} & $n$=\textbf{20} & $n$=\textbf{40} & $n$=\textbf{80} \\ \midrule
%Amazon       & 1.2988  & 1.3013   & 1.3019   \\
Amazon-movie & 1.1827  & 1.1842   & 1.1847   \\
Googlelocal  & 1.0995  & 1.1007   & 1.1013   \\
TripAdvisor  & 1.1299  & 1.1301   & 1.1305   \\
Yelp         & 1.0037  & 1.0050   & 1.0058   \\ \bottomrule
\end{tabular}
\label{tb:RatPred:Impn}
\end{table}

\noindent
{\bf Impact of arrival order. } 
The are two types of weights on ratings, 
i.e., based on arrival order and based on arrival time stamp of ratings.  
Here we study the weight which is based on arrival order.  
We fix $n=20$.   
To study the impact of arrival order, 
we consider three types of weights associated with the arrival order of ratings, 
i.e., exponential, polynomial and logarithmic in the arrival order, 
which are stated in Table \ref{tb:RatPred:ImpOrder}.   
Table \ref{tb:RatPred:ImpOrder} shows that RMSE of our R-BE method 
under these three types of weights.  
Consider that the weight $\eta_{k,j}$ is exponential in the arrival order, i.e., $\eta_{k,j} = \exp ( - \zeta j)$.  
One can observe that as we vary the parameter $\zeta$ of $\eta_{k,j} = \exp ( - \zeta j)$ 
from 0.01 to 0.0001,  
the RMSE can be further reduced over the unweighted case, i.e., $\zeta=0$.  
Similar observations can be found 
when the weight $\eta_{k,j}$ is polynomial or logarithmic in the arrival order of ratings.   
This implies that one can further improve the rating prediction accuracy 
via tuning the weight of ratings based on the arrival order.  
This improvement on rating prediction accuracy is supported by 
that users tend to assign larger weights to more recent ratings \cite{}
Furthermore, this improvement 
is achieved at simple selections of the weight of ratings.  
One can further improve the rating prediction accuracy by finer tuning of weights.   

\begin{table}[htb]
 \caption{\bf Impact of arrival order on rating prediction ($n{=}20$).}
\begin{tabular}{p{30 pt}p{30 pt}p{30 pt}||p{30 pt}p{30 pt}p{30 pt}}
\toprule
& & & \multicolumn{3}{c}{$\eta_{k,j} = \exp ( - \zeta j )$} \\ \midrule
\multicolumn{2}{c}{\textbf{category}} &Herd & $\zeta$\textbf{=0.01} & $\zeta$\textbf{=0.001} & $\zeta$\textbf{=0.0001} \\ \midrule
%\multicolumn{2}{c}{Amazon}       & 1.2988 & 1.3005 & 1.2994 &1.2991 \\
\multicolumn{2}{l}{Amazon-movie} & 1.4871 & \textbf{1.1823}  & \textbf{1.1824} & \textbf{1.1824} \\
\multicolumn{2}{l}{Googlelocal}  & 1.1412 & \textbf{1.0999} & \textbf{1.0999} & \textbf{1.0999} \\
\multicolumn{2}{l}{TripAdvisor}  & 1.1838 & \textbf{1.1300} & \textbf{1.1320} & \textbf{1.1311} \\
\multicolumn{2}{l}{Yelp}         & 1.4200 & \textbf{1.0033} & \textbf{1.0033} & \textbf{1.0033} \\ \midrule \midrule
\multicolumn{3}{c}{$\eta_{k,j} = j ^ {- \zeta}$} ||& \multicolumn{3}{c}{$\eta_{k,j} = 1 / [ \zeta ln ( j + 1 ) ]$}\\ \midrule
$\zeta$\textbf{=0.5} & $\zeta$\textbf{=1} & $\zeta$\textbf{=2}  & $\zeta$\textbf{=0.5} & $\zeta$\textbf{=1} & $\zeta$\textbf{=2} \\ \midrule
%\textbf{1.2970} & 1.3057 & 1.3037  & \textbf{1.2951}  & \textbf{1.2949} & \textbf{1.2963} \\
\textbf{1.1817} & \textbf{1.1818} & \textbf{1.1818}  & \textbf{1.1817} & \textbf{1.1814} & \textbf{1.1831} \\
\textbf{1.0993} & \textbf{1.0993} & \textbf{1.0993} & \textbf{1.0993} & \textbf{1.1001} & \textbf{1.1026} \\
 \textbf{1.1299} & \textbf{1.1299} & \textbf{1.1299}  & \textbf{1.1299} & \textbf{1.1303} & \textbf{1.1287}  \\
\textbf{1.0028} & \textbf{1.0028} & \textbf{1.0028}  & \textbf{1.0028} & \textbf{1.0043} & \textbf{1.0038} \\ \bottomrule
\end{tabular}
\label{tb:RatPred:ImpOrder}
\end{table}

\noindent 
{\bf Impact of arrival time stamp. }  
We fix $n=20$.   
To study the impact of arrival time stamp, 
we consider three types of weights associated with the arrival time stamp of ratings, 
i.e., exponential, polynomial and logarithmic in the arrival time stamp, 
which are stated in Table \ref{tb:RatPred:Imptime}.   
Table \ref{tb:RatPred:Imptime} shows that RMSE of our R-BE method 
under these three types of weights.  
Consider that the weight $\eta_{k,j}$ is exponential in the arrival time stamp, 
i.e., $\eta_{k,j} = \exp ( - \rho ( t_u - t_{i,k+1-j} ) ^ {0.4 } )$.  
One can observe that as we vary the parameter 
$\rho$ of $\eta_{k,j} = \exp ( - \rho ( t_u - t_{i,k+1-j} ) ^ {0.4 } )$ 
from 0.01 to 0.0001,  
the RMSE can be further reduced over the unweighted case, i.e., $\rho=0$.  
Similar observations can be found 
when the weight $\eta_{k,j}$ is polynomial or logarithmic in the arrival time stamp of ratings.   
This implies that one can further improve the rating prediction accuracy 
via tuning the weight of ratings based on the arrival time stamp.  
This improvement on rating prediction accuracy is supported by 
that users tend to assign larger weights to more recent ratings \cite{Rudolph2015}.
Furthermore, this improvement 
is achieved at simple selections of the weight of ratings.  
It can be further improved by finer tuning of weights.

\begin{table}[htb]
 \caption{\bf Imp. of arrival time stamp on rat. prediction ($n{=}20$).}
\begin{tabular}{p{30 pt}p{30 pt}p{30 pt}||p{30 pt}p{30 pt}p{30 pt}}
\toprule
& & & \multicolumn{3}{c}{$\eta_{k,j} = \exp ( - \rho ( t_u - t_{i,k+1-j} ) ^ {0.4 } )$ } \\ \midrule
\multicolumn{2}{c}{\textbf{category}} Herd & $\rho$\textbf{=0.01} & $\rho$\textbf{=0.001} & $\rho$\textbf{=0.0001} \\ \midrule
%\multicolumn{2}{c}{Amazon}       & 1.2988 & 1.3059 & 1.3001 & 1.3004 \\
\multicolumn{2}{l}{Amazon-movie} & 1.4871 & \textbf{1.1823}  & \textbf{1.1823} & \textbf{1.1823} \\
\multicolumn{2}{l}{Googlelocal}  & 1.1412 & \textbf{1.0998} & \textbf{1.0999} & \textbf{1.0999} \\
\multicolumn{2}{l}{TripAdvisor}  & 1.1838 & \textbf{1.1300} & \textbf{1.1300} & \textbf{1.1300} \\
\multicolumn{2}{l}{Yelp}         & 1.4200 & \textbf{1.0037} & \textbf{1.0037} & \textbf{1.0037} \\ \midrule \midrule
\multicolumn{3}{c}{$\eta_{k,j} = (  t_u - t_{i,k+1-j} ) ^ {- \rho}$} ||& \multicolumn{3}{c}{$\eta_{k,j} = 1 / \rho ln ( t_u - t_{i,k+1-j} + 1 )$} \\ \midrule
$\rho$\textbf{=0.2} & $\rho$\textbf{=0.4} & $\rho$\textbf{=0.8} & $\rho$\textbf{=0.2} & $\rho$\textbf{=0.4} & $\rho$\textbf{=0.8}\\ \midrule
%1.3002 & \textbf{1.2985} & \textbf{1.2987} & \textbf{1.2987} & 1.3009 & 1.3002 \\
\textbf{1.1623} & \textbf{1.1623} & \textbf{1.1623}  & \textbf{1.1623} & \textbf{1.1623} & \textbf{1.1623} \\
\textbf{1.0991} & \textbf{1.0991} & \textbf{1.0999} & \textbf{1.0992} & \textbf{1.00989} & \textbf{1.0999} \\
 \textbf{1.1299} & \textbf{1.1299} & \textbf{1.1303}  & \textbf{1.1320} & \textbf{1.1320} & \textbf{1.1320}  \\
\textbf{1.0031} & \textbf{1.0031} & \textbf{1.0049}  & \textbf{1.0031} & \textbf{1.0030} & \textbf{1.0045} \\ \bottomrule
\end{tabular}
\label{tb:RatPred:Imptime}
\end{table}

\subsection{Aggregate Opinion Heterogeneity for Rating Prediction}

In this section, we study the benefit of exploiting aggregate opinion
heterogeneity for rating prediction. 
We consider the initial opinion formation model derived in 
Equation (\ref{eq:Model:IniOpOrder}).  
Table \ref{tb:RatPred:ImpAggopExp} shows the RMSE of our R-BE method 
under the case that the weight $\beta_m$ is exponential in opinion levels.   
In Table  \ref{tb:RatPred:ImpAggopExp}, we only compare our R-BE method  with Herd.  
One can observe that as we vary the parameter $\gamma$ of $\beta_m = \exp ( - \gamma m )$  
from 0.01 to 0.0001,  
the RMSE can be further reduced over Herd.  
Similar observations can be found 
when the weights $\beta_m$ is polynomial or logarithmic in rating level $m$ 
as shown in Table \ref{tb:RatPred:ImpAggopply} and \ref{tb:RatPred:ImpAggoplog}.    
This implies that one can further improve the rating prediction accuracy 
via tuning the weight of rating levels, i.e., aggregate opinion heterogeneity.    
This improvement on rating prediction accuracy is supported by 
that users tend to assign different weights to different rating levels \cite{Rudolph2015}
This reduction of RMSE is achieved at simple selections on weight for rating levels.  
Finer tuning of weight may lead to further reduction on the RMSE.

% \usepackage{booktabs}
\begin{table}[htb]
 \caption{\bf Exponential weights in opinion level for rat. pred.  }
\begin{tabular}{p{60 pt}||p{35 pt}p{35 pt}p{35 pt}p{35 pt}}
\toprule
& \multicolumn{4}{c}{$\beta_m = \exp ( - \gamma m )$}\\ \midrule
\textbf{category} & \textbf{Herd} & $\gamma$=\textbf{0.1} & $\gamma$=\textbf{0.01} &$ \gamma$\textbf{=0.001} \\ \midrule
%Amazon       & 1.3134 & \textbf{1.3095} & \textbf{1.3102} & \textbf{1.3103} \\
Amazon-movie & 1.4871 & \textbf{1.1661} &  \textbf{1.1824} & \textbf{1.1844} \\
Googlelocal  & 1.1412 & \textbf{1.1155} & \textbf{1.1026} & \textbf{1.1016} \\
TripAdvisor  & 1.1838 & \textbf{1.1387} & \textbf{1.1312} & \textbf{1.1308} \\
Yelp         & 1.4200 & \textbf{0.9929} & \textbf{1.0043} & \textbf{1.0057} \\ \midrule
\textbf{category} & \textbf{Herd} & $\gamma$=\textbf{-0.1} & $\gamma$=\textbf{-0.01} & $\gamma$=\textbf{-0.001} \\ \midrule
%Amazon       & 1.3134 & 1.3160 & \textbf{1.3103} & \textbf{1.3057} \\
Amazon-movie & 1.4871 & \textbf{1.2115} &  \textbf{1.1871} & \textbf{1.1851} \\
Googlelocal  & 1.1412 & \textbf{1.0923} & \textbf{1.1002} & \textbf{1.1018} \\
TripAdvisor  & 1.1838 & \textbf{1.1292} & \textbf{1.1303} & \textbf{1.1307} \\
Yelp         & 1.4200 & \textbf{1.0230} & \textbf{1.0074} & \textbf{1.0062} \\ \bottomrule
\end{tabular}
\label{tb:RatPred:ImpAggopExp}
\end{table}
 
% \usepackage{booktabs}
\begin{table}[htb]
 \caption{\bf Polynomial weights in opinion level for rat. pred. }
\begin{tabular}{p{60 pt}||p{35 pt}p{35 pt}p{35 pt}p{35 pt}}
\toprule
& \multicolumn{4}{c}{$\beta_m = m^{\gamma}$}\\ \midrule
\textbf{category} & \textbf{Herd} & $\gamma$\textbf{=0.5} & $\gamma$\textbf{=1} & $\gamma$\textbf{=2} \\ \midrule
%Amazon       & 1.3134 & \textbf{1.3093} & \textbf{1.3107} & \textbf{1.3049} \\
Amazon-movie & 1.4871 & \textbf{1.1618} &  \textbf{1.1613} & \textbf{1.1615} \\
Googlelocal  & 1.1412 & \textbf{1.1203} & \textbf{1.1449} & \textbf{1.1436} \\
TripAdvisor  & 1.1838 & \textbf{1.1437} & \textbf{1.1690} & \textbf{1.1463} \\
Yelp         & 1.4200 & \textbf{0.9896} & \textbf{0.9846} & \textbf{0.9833} \\ \midrule
\textbf{category} & \textbf{Herd} & $\gamma$\textbf{=-0.5} & $\gamma$\textbf{=-1} & $\gamma$\textbf{=-2} \\ \midrule
%Amazon       & 1.3134 & \textbf{1.3104} & 1.3143 & \textbf{1.3100} \\
Amazon-movie & 1.4871 & \textbf{1.2149} &  \textbf{1.2388} & \textbf{1.2621} \\
Googlelocal  & 1.1412 & \textbf{1.0926} & \textbf{1.0889} & \textbf{1.0868} \\
TripAdvisor  & 1.1838 & \textbf{1.1293} & \textbf{1.1310} & \textbf{1.1340} \\
Yelp         & 1.4200 & \textbf{1.0229} & \textbf{1.0359} & \textbf{1.0498} \\ \bottomrule
\end{tabular}
\label{tb:RatPred:ImpAggopply}
\end{table}

% \usepackage{booktabs}
\begin{table}[htb]
 \caption{\bf Logarithmic weights in opinion level for rat. pred.}
\begin{tabular}{p{60 pt}||p{35 pt}p{35 pt}p{35 pt}p{35 pt}}
\toprule
& \multicolumn{4}{c}{$\beta_m = ln ( m^{\gamma} + 1 )$}\\ \midrule
\textbf{category} & \textbf{Herd} & $\gamma$\textbf{=0.5} & $\gamma$\textbf{=1} & $\gamma$\textbf{=2} \\ \midrule
%Amazon       & 1.3134 & \textbf{1.3062} & \textbf{1.3020} & \textbf{1.3084} \\
Amazon-movie & 1.4871 & \textbf{1.1608} &  \textbf{1.1607} & \textbf{1.1619} \\
Googlelocal  & 1.1412 & \textbf{1.1261} & \textbf{1.1480} & \textbf{1.1376} \\
TripAdvisor  & 1.1838 & \textbf{1.1495} & \textbf{1.1705} & \textbf{1.1482} \\
Yelp         & 1.4200 & \textbf{0.9871} & \textbf{0.9839} & \textbf{0.9846} \\ \midrule
\textbf{category} & \textbf{Herd} & $\gamma$\textbf{=-0.5} & $\gamma$\textbf{=-1} & $\gamma$\textbf{=-2} \\ \midrule
%Amazon       & 1.3134 & \textbf{1.3080} & \textbf{1.3076} & \textbf{1.3107} \\
Amazon-movie & 1.4871 & \textbf{1.2331} &  \textbf{1.2729} & \textbf{1.2842} \\
Googlelocal  & 1.1412 & \textbf{1.0893} & \textbf{1.0858} & \textbf{1.0876} \\
TripAdvisor  & 1.1838 & \textbf{1.1305} & \textbf{1.1370} & \textbf{1.1474} \\
Yelp         & 1.4200 & \textbf{1.0336} & \textbf{1.0620} & \textbf{1.0941} \\ \bottomrule
\end{tabular}
\label{tb:RatPred:ImpAggoplog}
\end{table}

\section{\bf Related Work}
\label{sec:relatedwork}

Online product rating (or review) systems has been studied extensively.
A number of works investigated whether and how online product rating systems
can benefit sellers and users.
Chevalier {\em et al.} \cite{Chevalier2006} studied the impact of
product reviews on the sales of sellers, and they found that
positive product reviews can increase the sales.
Mudambi {\em et al.} \cite{Mudambi2010} studied the impact of
product reviews on customer's purchasing behavior,
and found that product reviews are helpful in purchasing decision makings.
Similar observations were found by
Lackermair {\em et al.} \cite{Lackermair2013} and Li \cite{Li2013}.
We refer readers to \cite{BrightLocal2016,Rudolph2015,Shrestha2016}
for a number of survey studies on the role of product reviews in
purchasing decisions.

A variety of works investigated rating (or review) biases.
A number of sources that can lead to rating biases has been revealed,
e.g., product categories \cite{Guo2015},
system interfaces \cite{Cosley2003},
recommendation algorithms \cite{Shafto2016},
the dynamics of user preferences \cite{Koren2009},
the improvement of user expertise \cite{McAuley2013}, etc.
To mitigate these biases, a number of methods or
algorithms were developed,
e.g., \cite{Cosley2003,Guo2015,Koren2009,McAuley2013,Shafto2016}.
Our work is closed related to the studied of rating bias
from psychological perspectives.
Zhang {\em et al.} \cite{Zhang2017} modeled the assimilate
and contrast phenomenon in ratings,
and they used the model to improve recommendation accuracy.
The herding effects in product ratings were
revealed by a number of real-workd experiments
\cite{Salganik2006,Muchnik2013}.
Krishnan {\em et al.} \cite{Krishnan2014}
and Wang {\em et al.} \cite{Wang2014}
developed models to quantify the effect of herding effects
(or social influence biases).  
Coba {\em et al.} \cite{Coba2018} did some experiments to 
study how rating summaries influence user decisions.  
These works provided evidences for the existence of herding effects
in online product rating systems.
Built on these evidences, 
we investigate the convergence of product ratings.  
We also demonstrate how to utilize these convergence properties to 
manage product ratings.

\section{\bf Conclusion}
\label{sec:conclusion}

We develop a framework to manage online product ratings.
%so as to eliminate the rating bias cause by the herding effects.
We formulate a mathematical model to characterize the herding effects,
and the decision space to correct product ratings.
We identify a class of rating aggregation rules,
under which the historical collective opinion
converges to the ground-truth collective opinion.
We derive a metric to quantify the speed of convergence,
which also guides product rating managing.  
%We prove that the herding effects slows down the
%speed of convergence and an accurate
%review selection mechanism can speed up it.
%This metric also quantifies the
%speed of convergence for each rating aggregation rule.
%We conduct experiments on synthetic data and real data. 
Via theoretical analysis and experiment studies we found:
(1) recency aware aggregation rules
can significantly speed up the convergence over the unweighted average rule
(commonly deployed) especially under strong herding effects;
(2) the convergence speed increases in the accuracy of the review selection mechanism but this improvement becomes small when the aggregation rule
has a strong recency awareness;
(3) recency aware rating aggregation rules can
improve the convergence speed
in Amazon and TripAdvisor by 41\% and 62\% respectively.

\bibliographystyle{IEEEtran}
\bibliography{Reference}

\appendix

\section{\bf Technical Proofs}
%\section*{}
\noindent
{\bf Proof of Lemma \ref{lem:ImpEta}: }
First we have
$
\bm{\theta}_i - \bm{\alpha}
= (1 - \eta_i) {\bm \beta}_i + \eta_i {\bm \alpha}
- {\bm \alpha}
= (1 - \eta_i) ({\bm \beta}_i - {\bm \alpha}).
$
Then if follows that
$
||\bm{\theta}_i - \bm{\alpha}||
= (1 - \eta_i)
||{\bm \beta}_i - {\bm \alpha}||.
$
Therefore, $||\bm{\theta}_i - \bm{\alpha}||$
is decreasing in $\eta_i$.
\done

\noindent
{\bf Proof of Theorem \ref{thm:convergence}: }
Let $\mathbf{e}_{\ell}
\triangleq
[e_{\ell,1}, \ldots, e_{\ell,M}] \in \mathbb{R}^M$
denote an $M$ dimensional vector,
whose $\ell$-th entry is 1, i.e., $e_{\ell,\ell} = 1$ and all the other entries are zero, i.e., $e_{\ell,m} = 0, \forall m \neq \ell$ and $m \in \mathcal{M}$.
Note that $\mathbf{I}_{\{R_j=m\}}$ equals to $e_{R_j, m}$.
Then we have that
\begin{align*}
{\bm \beta}_i
&
=
\left[
\frac{\sum^i_{j=1} w_j \mathbf{I}_{\{R_j = 1\}} }
{\sum^i_{j=1} w_j},
\ldots,
\frac{\sum^i_{j=1} w_j \mathbf{I}_{\{R_j = M\}} }
{\sum^i_{j=1} w_j}
\right]
\\
&
=
\left[
\frac{\sum^i_{j=1} w_j e_{R_j, 1} }
{\sum^i_{j=1} w_j},
\ldots,
\frac{\sum^i_{j=1} w_j e_{R_j, M} }
{\sum^i_{j=1} w_j}
\right]
\\
&= \frac{\sum^i_{j=1} w_j \mathbf{e}_{R_j} }
{\sum^i_{j=1} w_j}.
\end{align*}
This implies a useful equation for later proof
$
\sum^i_{j=1} w_j \mathbf{e}_{R_j}
= {\bm \beta}_i  \sum^i_{j=1} w_j.
$
Then we can derive  $\bm{\beta}_{i+1}$ as follows:
\begin{align*}
\bm{\beta}_{i+1}
&
=
\frac{\sum^{i+1}_{j=1} w_j \mathbf{e}_{R_j} }
{\sum^{i+1}_{j=1} w_j}
= \frac{w_{i+1} \mathbf{e}_{R_{i+1}}
+ \sum^{i}_{j=1} w_j \mathbf{e}_{R_j} }
{\sum^{i+1}_{j=1} w_j}
\\
&
=
\frac{w_{i+1} \mathbf{e}_{R_{i+1}}
+ \bm{\beta}_{i}  \sum^{i}_{j=1} w_j }
{\sum^{i+1}_{j=1} w_j}
\\
&
=
\left(1 - \frac{w_{i+1}}{\sum^{i+1}_{j=1} w_j} \right)
\bm{\beta}_{i}
+
\frac{w_{i+1}}{\sum^{i+1}_{j=1} w_j}
\mathbf{e}_{R_{i+1}}
\\
& =
\left(1 - \tilde{w}_{i+1} \right)
\bm{\beta}_{i}
+
\tilde{w}_{i+1}
\left[
\gamma_i \bm{\theta}_{i} +
(1 - \gamma_i) \bm{\alpha}
+ \mathbf{E}_{i+1}
\right],
\end{align*}
where we define $\mathbf{E}_{i+1}$ as
$
\mathbf{E}_{i+1}
\triangleq
\mathbf{e}_{R_{i+1}}
- \gamma_i \bm{\theta}_{i}
- (1 - \gamma_i) \bm{\alpha} .
$
Let $H_i$ denote a mapping defined as
$
H_i \bm{\beta}_i
\triangleq
\gamma_i \bm{\theta}_i
+ (1 - \gamma_i) \bm{\alpha}.
$
Then it follows that $H_i$ is a pseudo-contraction mapping
\[
\| H_i \bm{\beta}_i - \bm{\alpha} \|
= \gamma_i \| \bm{\theta}_i - \bm{\alpha} \|
\leq \gamma_i \| \bm{\beta}_i - \bm{\alpha} \|
\leq \sup_{i \in \mathbb{N}_+} \gamma_i
\| \bm{\beta}_i - \bm{\alpha} \|.
\]
Note that we conclude the pseudo-contraction mapping,
because $\sup_{i \in \mathbb{N}_+} \gamma_i  <1$.
Then it follows that $H_i$ has a unique fixed point $\bm{\alpha}$.
One can also check that the error vector $\mathbf{E}_{i+1}$ satisfies
\begin{align*}
\mathbb{E} [\mathbf{E}_{i+1} | \mathcal{H}_i]
&
=
\mathbb{E} [\mathbf{e}_{R_{i+1}} | \mathcal{H}_i]
- \gamma_i \bm{\theta}_{i}
- (1 - \gamma_i ) \bm{\alpha}
\\
&
= \gamma_i \bm{\theta}_i + (1 - \gamma_i) \bm{\alpha}
- \gamma_i \bm{\beta}_{i}
- (1 - \gamma_i ) \bm{\alpha}
= \mathbf{0}.
\end{align*}
Furthermore,
$
\mathbb{E} [E^2_{i+1, m} | \mathcal{H}_i]
\leq [\gamma_i \theta_{i,m}
+ (1 - \gamma_i ) \alpha_m]^2
+ \mathbb{E}[\mathbf{I}^2_{\{R_{i+1}=m\}} | \mathcal{H}_i]
\leq 2.
$
Lastly, note that $\tilde{w}_i$ satisfies
$
\sum^\infty_{i=1}
\tilde{w}_i
= \infty,
\sum^\infty_{i=1}
\tilde{w}^2_i
< \infty.
$
Then it follows that $\bm{\beta}_i$
converges to the unique fixed point of $H_i$ almost surely.
Then it follows that
$
\mathbb{P}
[\lim_{i \rightarrow \infty}
\bm{\beta}_i = \bm{\alpha}
] = 1.
$
\done

\noindent
{\bf Proof of Theorem \ref{thm:ImpMisbConv}: }
Note that the number of misbehaving ratings is finite
and the index of the last misbehaving rating is $i_k$.
Let us now consider $\bm{\beta}_i$ for all
$i = i_{k}+1, \ldots, \infty$.
Note that from the proof of Theorem \ref{thm:convergence},
one can observe that the convergence of $\bm{\beta}_i$
is invariant of the initial historical collective opinion,
i.e., $\bm{\beta}_1$.
Therefore the sequence of historical collective opinion
$\bm{\beta}_{i_k}, \bm{\beta}_{i_k + 1}, \ldots,
\infty$ still converges to $\bm{\alpha}$ almost surely.
\done

\noindent
{\bf Proof of Theorem \ref{thm:ConvRate}: }
Without loss of generality,
let us consider one entry of the historical collective opinion $\bm{\beta}_i$,
i.e., $\beta_{i,m}$.
From the proof of Theorem \ref{thm:convergence}, we have that
$
\beta_{i+1,m}
=
(1 - \tilde{w}_{i+1}) \beta_{i,m}
+ \tilde{w}_{i+1} \mathbf{I}_{\{R_{i+1} = m\}}.
$
Then it follows that
\begin{align*}
&
\mathbb{E} [\beta_{i+1,m} | \mathcal{H}_i]
\\
&= (1 - \tilde{w}_{i+1}) \beta_{i,m}
+ \tilde{w}_{i+1} [ \gamma_i \theta_{i,m} + (1-\gamma_i) \alpha_m]
\\
&
= (1 - \tilde{w}_{i+1}) \beta_{i,m}
+ \tilde{w}_{i+1} [\gamma_i
((1 - \eta_i) \beta_{i,m} + \eta_i \alpha_{i,m})
\\
& +  (1-\gamma_i) \alpha_m]
\\
&
= (1 - \tilde{w}_{i+1}) \beta_{i,m}
+ \tilde{w}_{i+1} [\gamma_i (1 - \eta_i) \beta_{i,m}
\\
& + (\eta_i  \gamma_i +1 - \gamma_i )  \alpha_{i,m}]
\\
&
=
\beta_{i,m} + \tilde{w}_{i+1}
(\eta_i  \gamma_i +1 - \gamma_i)
(\alpha_m - \beta_{i,m})
\end{align*}
Let $\tilde{\beta}_{i,m} = \beta_{i,m} - \alpha_m$.
Then we can further have
\begin{align*}
& \mathbb{E} [\tilde{\beta}_{i+1,m} | \mathcal{H}_i]
= \mathbb{E} [\beta_{i+1,m} - \alpha_m | \mathcal{H}_i]
\\
&
= \beta_{i,m} + \tilde{w}_{i+1}
(\eta_i  \gamma_i +1 - \gamma_i)
(\alpha_m - \beta_{i,m}) - \alpha_m
\\
&
= \tilde{\beta}_{i,m} - \tilde{w}_{i+1}
(\eta_i  \gamma_i +1 - \gamma_i) \tilde{\beta}_{i,m}
\\
&
= [1 - \tilde{w}_{i+1} (\eta_i  \gamma_i +1 - \gamma_i) ]
\tilde{\beta}_{i,m}.
\end{align*}
Note that
$
\varphi_j (\mathbf{w}, \bm{\beta}, \bm{\gamma})
\triangleq
\prod^j_{\ell=1}
[1 - \tilde{w}_{\ell+1} (\eta_{\ell}  \gamma_{\ell} +1 - \gamma_{\ell}) ],
$
and define $\varphi_0 (\mathbf{w}, \bm{\beta}, \bm{\gamma}) = 1$.
For the simplicity of presentation, let us define
$
\hat{\beta}_{i,m}
\triangleq
\frac{\tilde{\beta}_{i,m}}
{\varphi_{i-1} (\mathbf{w}, \bm{\beta}, \bm{\gamma})}.
$
Then it follows that
\begin{align*}
\mathbb{E} [\hat{\beta}_{i+1,m} | \mathcal{H}_i]
&
= \mathbb{E}
\left[
\frac{\beta_{i+1,m}}
  { \varphi_{i}(\mathbf{w}, \bm{\beta}, \bm{\gamma}) }
| \mathcal{H}_i \right]
= \frac{\mathbb{E}
\left[\beta_{i+1,m} | \mathcal{H}_i \right] }
{\varphi_{i} (\mathbf{w}, \bm{\beta}, \bm{\gamma}) }
\\
& \hspace{-0.38 in}
= \frac{1 - \tilde{w}_{i+1} (\eta_i  \gamma_i +1 - \gamma_i) }
{\varphi_{i}  (\mathbf{w}, \bm{\beta}, \bm{\gamma}) }
\tilde{\beta}_{i,m}
\\
&
= \frac{\tilde{\beta}_{i,m}}
{\varphi_{i-1} (\mathbf{w}, \bm{\beta}, \bm{\gamma})  }
 = \hat{\beta}_{i,m}.
\end{align*}
Note that
$
\mathbb{E}[|\hat{\beta}_{i,m}|]
\leq \frac{1}{\varphi_{i-1} (\mathbf{w}, \bm{\beta}, \bm{\gamma})  } < \infty.
$
Thus $\hat{\beta}_{i,m}$ forms a martingale with respect to
$\mathcal{H}_i$.
To study the convergence rate of $\hat{\beta}_{i,m}$,
let us first check the difference between
$\hat{\beta}_{i+1,m}$ and $\hat{\beta}_{i,m}$.
\begin{align*}
&  
\hat{\beta}_{i+1,m} - \hat{\beta}_{i,m}
\\
& 
=
\frac{\tilde{\beta}_{i+1,m}}
  {\varphi_{i} (\mathbf{w}, \bm{\beta}, \bm{\gamma})  }
-
\frac{\tilde{\beta}_{i,m}}
  {\varphi_{i-1} (\mathbf{w}, \bm{\beta}, \bm{\gamma})  }
\\
&
=
\frac{\tilde{\beta}_{i+1,m}
- [1 - \tilde{w}_{i+1} (\eta_i  \gamma_i +1 - \gamma_i) ]
\tilde{\beta}_{i,m} }
{\varphi_{i}  (\mathbf{w}, \bm{\beta}, \bm{\gamma}) }
\\
&
=
\frac{\tilde{\beta}_{i+1,m} - \tilde{\beta}_{i,m}
+ \tilde{w}_{i+1} (\eta_i  \gamma_i +1 - \gamma_i)
\tilde{\beta}_{i,m} }
{\varphi_{i}  (\mathbf{w}, \bm{\beta}, \bm{\gamma}) }
\\
&
=
\frac{\beta_{i+1,m} - \beta_{i,m}
+\tilde{w}_{i+1} (\eta_i  \gamma_i +1 - \gamma_i)
\tilde{\beta}_{i,m} }
{\varphi_{i}  (\mathbf{w}, \bm{\beta}, \bm{\gamma}) }
\\
&
=
\frac{\tilde{w}_{i+1} (\mathbf{I}_{\{R_{i+1} = m\}} - \beta_{i,m})
+ \tilde{w}_{i+1} (\eta_i  \gamma_i +1 - \gamma_i)
 \tilde{\beta}_{i,m} }
{\varphi_{i}  (\mathbf{w}, \bm{\beta}, \bm{\gamma}) }
\\
&
= \frac{\tilde{w}_{i+1}}
  {\varphi_{i}  (\mathbf{w}, \bm{\beta}, \bm{\gamma}) }
\left[
\mathbf{I}_{\{R_{i+1} = m\}} - \beta_{i,m}
+ (\eta_i  \gamma_i +1 - \gamma_i)  \tilde{\beta}_{i,m}
\right].
\end{align*}
For simplicity of presentation, we define
\begin{align*}
&
b_{i+1}
\triangleq
\frac{\tilde{w}_{i+1}}{\varphi_{i} (\mathbf{w}, \bm{\beta}, \bm{\gamma}) }
\left[
1 - \beta_{i,m}
+ (\eta_i  \gamma_i +1 - \gamma_i)  \tilde{\beta}_{i,m}
\right],
\\
&
a_{i+1}
\triangleq
\frac{\tilde{w}_{i+1}}{\varphi_{i} (\mathbf{w}, \bm{\beta}, \bm{\gamma}) }
\left[ (\eta_i  \gamma_i +1 - \gamma_i)  \tilde{\beta}_{i,m}
- \beta_{i,m}
\right],
\end{align*}
where $i = 0, 1, \dots, \infty$. Then it follows that
$
a_{i+1}
\leq
\hat{\beta}_{i+1,m} - \hat{\beta}_{i,m}
\leq
b_{i+1}
$
Furthermore, we we have
$
b_i - a_i = \frac{\tilde{w}_{i}}
  {\varphi_{i-1} (\mathbf{w}, \bm{\beta}, \bm{\gamma}) },
\forall i \in \mathbb{N}_+.
$
Note that $\bm{\theta}_0 = \bm{\alpha}$.
This implies that $\mathbb{E}[\hat{\beta}_{1,m}] = 0$.
Then we have $\mathbb{E}[\hat{\beta}_{i,m}]
= \mathbb{E}[\hat{\beta}_{1,m}] = 0$.
Then applying the martingale concentration theory
(in particular the Azuma Hoeffiding inequality) we have
\begin{align*}
&
\mathbb{P} [|\beta_{i,m} - \alpha_m| > \epsilon]
= \mathbb{P}
\left[ \frac{|\beta_{i,m} - \alpha_m|}
  {\varphi_{i-1} (\mathbf{w}, \bm{\beta}, \bm{\gamma})}
>
\frac{\epsilon}
  {\varphi_{i-1} (\mathbf{w}, \bm{\beta}, \bm{\gamma})}
\right]
\\
&
= \mathbb{P}
\left[ |\hat{\beta}_{i,m}| >
\frac{\epsilon}
  {\varphi_{i-1} (\mathbf{w}, \bm{\beta}, \bm{\gamma}) } \right]
\\
&
\leq
2 \exp \left(
- \frac{2 \epsilon^2}
{\varphi^2_{i-1} (\mathbf{w}, \bm{\beta}, \bm{\gamma})}
\frac{1}{\sum^i_{j=1} (b_i-a_i)^2 }
\right)
\\
&
= 2 \exp \left(
- \frac{2}
{\varphi^2_{i-1} (\mathbf{w}, \bm{\beta}, \bm{\gamma})}
\frac{1}{\sum^i_{j=1}
 \frac{\tilde{w}^2_{j}}
  {\varphi^2_{j-1} (\mathbf{w}, \bm{\beta}, \bm{\gamma}) }
}
\epsilon^2
\right)
\\
&
= 2 \exp(-\phi_i (\mathbf{w}, \bm{\beta}, \bm{\gamma}) \epsilon^2),
\end{align*}
This proof is then complete.
\done

{\bf Proof of Theorem \ref{thm:ImpGamEtaSpeed}: }
First we can rewrite $\phi_i (\mathbf{w}, \bm{\beta}, \bm{\gamma}) $ as
\begin{align*}
\phi_i (\mathbf{w}, \bm{\beta}, \bm{\gamma})
&
= 1 \bigg/
\left(
\frac{\varphi^2_{i-1} (\mathbf{w}, \bm{\eta}, \bm{\gamma}) }{2}
\sum^i_{j=1}
\frac{\tilde{w}^2_j }
{\varphi^2_{j-1} (\mathbf{w}, \bm{\eta}, \bm{\gamma})} \right)
\\
&= 1 \bigg/
\left(
\frac{1}{2}
\sum^i_{j=1}
\left(
\frac{\varphi_{i-1} (\mathbf{w}, \bm{\eta}, \bm{\gamma}) }
{\varphi_{j-1} (\mathbf{w}, \bm{\eta}, \bm{\gamma})}
\right)^2
\tilde{w}^2_j
\right).
\end{align*}
Note that we can further have
\begin{align*}
\frac{\varphi_{i-1} (\mathbf{w}, \bm{\eta}, \bm{\gamma}) }
{\varphi_{j-1} (\mathbf{w}, \bm{\eta}, \bm{\gamma})}
&
= \frac{\prod^{i-1}_{\ell=1}
\left[
1 - \tilde{w}_{\ell+1} (1 - \gamma_{\ell} + \eta_{\ell}  \gamma_{\ell} )
\right]}
{
\prod^{j-1}_{\ell=1}
\left[
1 - \tilde{w}_{\ell+1} (1 - \gamma_{\ell} + \eta_{\ell}  \gamma_{\ell} )
\right]}
\\
&
= \prod^{i-1}_{j}
\left[
1 - \tilde{w}_{\ell+1} (1 - \gamma_{\ell} + \eta_{\ell}  \gamma_{\ell} )
\right].
\end{align*}
Observe that the term
$
1 - \tilde{w}_{\ell+1} (1 - \gamma_{\ell} + \eta_{\ell}  \gamma_{\ell} )
=
1 - \tilde{w}_{\ell+1} (1 - (1- \eta_{\ell} ) \gamma_{\ell} ) $.
Thus the term
$1 - \tilde{w}_{\ell+1} (1 - \gamma_{\ell} + \eta_{\ell}  \gamma_{\ell} ) $
is non-decreasing in $\gamma_\ell$
and is non-increasing in $\eta_\ell$.
Therefore,
$
\varphi_{i-1} (\mathbf{w}, \bm{\eta}, \bm{\gamma})
/ \varphi_{j-1} (\mathbf{w}, \bm{\eta}, \bm{\gamma})
$
is non-decreasing in $\gamma_\ell$
and is non-increasing in $\eta_\ell$
for all $\ell \leq i$.
We can then conclude this theorem.
\done

\noindent
{\bf Proof of Theorem \ref{thm:ImpMisbConvRate}:}
Note that the number of misbehaving ratings is finite
and the index of the last misbehaving rating is $i_k$.
Let us now consider $\bm{\beta}_i$ for all
$i = i_{k}+1, \ldots, \infty$.
Let $\mu \triangleq
1/ \varphi_{i_k - 1} (\mathbf{w}, \bm{\eta}, \bm{\gamma})$.
Then it follows that
$
|\mathbb{E} [\hat{\beta}_{i_k, m}]| \leq
\frac{1}{\varphi_{i_k - 1}
 (\mathbf{w}, \bm{\eta}, \bm{\gamma}) }
= \mu.
$
Now we treat $\hat{\beta}_{i_k, m}$ as the starting point.
Note that the sequence
$\hat{\beta}_{i, m}, \forall i = i_k, \ldots, \infty$ forms a martingale,
i.e., $\mathbb{E} [\hat{\beta}_{i+1, m} | \mathcal{H}_i]
= \hat{\beta}_{i, m}$ for all $i = i_k, \ldots, \infty$.
Furthermore, the difference still holds
$
a_{i+1}
\leq
\hat{\beta}_{i+1,m} - \hat{\beta}_{i,m}
\leq
b_{i+1},
\forall i = i_k, \ldots, \infty.
$
and  $b_i, a_i$ satisfies
$
b_i - a_i = \frac{\tilde{w}_{i}}
{\varphi_{i-1} (\mathbf{w}, \bm{\eta}, \bm{\gamma}) },
\forall i = i_k+1, \ldots, \infty.
$
Note that $\mathbb{E}[\hat{\beta}_{i,m}]
= \mathbb{E}[\hat{\beta}_{i_k, m}] = \mu$
for all $i = i_k, \ldots, \infty$.
Then it follows that
\begin{align*}
&
\mathbb{P} [|\beta_{i,m} - \alpha_m| > \epsilon]
= \mathbb{P}
\left[ \frac{|\beta_{i,m} - \alpha_m|}{\varphi_{i-1}}
> \frac{\epsilon}
{\varphi_{i-1} (\mathbf{w}, \bm{\eta}, \bm{\gamma}) } \right]
\\
&
= \mathbb{P}
\left[ |\hat{\beta}_{i,m}| > \frac{\epsilon}
{\varphi_{i-1} (\mathbf{w}, \bm{\eta}, \bm{\gamma}) } \right]
\\
&
\leq
\mathbb{P}
\left[ |\hat{\beta}_{i,m} - \mathbb{E}[\beta_{i_k, m}]| >
\frac{\epsilon}
  {\varphi_{i-1} (\mathbf{w}, \bm{\eta}, \bm{\gamma}) }
-
|\mathbb{E}[\beta_{i_k, m}]|
\right]
\\
&
\leq
\mathbb{P}
\left[ |\hat{\beta}_{i,m} - \mathbb{E}[\beta_{i_k, m}]| >
\frac{\epsilon}
  {\varphi_{i-1} (\mathbf{w}, \bm{\eta}, \bm{\gamma}) }
-
\mu
\right]
\end{align*}
Then by some algebraic operations, we have
\begin{align*}
&
\mathbb{P} [|\beta_{i,m} - \alpha_m| > \epsilon]
\\
&
\leq
2 \exp \left(
- 2
\frac{(\epsilon - \mu \varphi_{i-1} (\mathbf{w}, \bm{\eta}, \bm{\gamma}))^2}
{\varphi^2_{i-1} (\mathbf{w}, \bm{\eta}, \bm{\gamma})}
\frac{\mathbf{I}_{\{\epsilon
\varphi_{i_k -1}(\mathbf{w}, \bm{\eta}, \bm{\gamma})
> \varphi_{i-1}(\mathbf{w}, \bm{\eta}, \bm{\gamma}) \}}}
{ \sum^i_{j=i_k + 1}
\frac{\tilde{w}^2_{j} }
{\varphi^2_{j-1} (\mathbf{w}, \bm{\eta}, \bm{\gamma})} }
\right)
\\
&
\leq
2 \exp \left(
- 2
\frac{(1 - \mu \varphi_{i-1} (\mathbf{w}, \bm{\eta}, \bm{\gamma})
/ \epsilon)^2}
{\varphi^2_{i-1} (\mathbf{w}, \bm{\eta}, \bm{\gamma})}
\frac{\mathbf{I}_{\{\epsilon
\varphi_{i_k -1}(\mathbf{w}, \bm{\eta}, \bm{\gamma})
> \varphi_{i-1}(\mathbf{w}, \bm{\eta}, \bm{\gamma}) \}}}
{\sum^i_{j=i_k + 1}
\frac{\tilde{w}^2_{j} }
{\varphi^2_{j-1} (\mathbf{w}, \bm{\eta}, \bm{\gamma})} }
\epsilon^2
\right)
\\
&
= 2 \exp(- \tilde{\phi}_i \epsilon^2).
\end{align*}
This proof is then complete.
\done

\noindent
{\bf Proof of Theorem \ref{cor:minNumRat}: }
Consider the average score rule.
Note that
$
i \geq
\phi^{-1}
\left(
\frac{M^2 (M+1)^2}{4 \epsilon^2}  \ln \frac{M}{2 \delta}
\right)
$
implies
$
\mathbb{P}
\left[|\beta_{i,m} - \alpha_m| > \frac{2 \epsilon}{M (M+1)}\right]
\leq \frac{\delta}{M}.
$
By the union bound, we have
\begin{align*}
& \hspace{-0.18 in}
\mathbb{P}
\left[
\exists m \in \mathcal{M},
|\beta_{i,m} - \alpha_m| > \frac{2 \epsilon}{M (M+1)}\right]
\\
&
\leq
\sum^M_{m=1}
\frac{\delta}{M}
\mathbb{P}
\left[|\beta_{i,m} - \alpha_m| > \frac{2 \epsilon}{M (M+1)}\right]
= \delta
\end{align*}
Namely, with probability at least $1-\delta$ we have that
$|\beta_{i,m} - \alpha_m| \leq \frac{2 \epsilon}{M (M+1)}$
holds for all $m \in \mathcal{M}$.
Then with probability at least $1-\delta$, it holds that
\begin{align*}
|A(\bm{\beta}_i) - A(\bm{\alpha})|
=
\left|
\sum^M_{m=1}
(\beta_{i,m} - \alpha_m)
\right|
\leq
\sum^M_{m=1}  \frac{2\epsilon}{M(M+1)}
= \epsilon.
\end{align*}
Now, let us consider the majority rule.
With a similar proof, one can conclude that
with probability at least $1-\delta$,
it holds that
$
|\beta_{i,m} - \alpha_m| <
\frac{\alpha_{max} - \alpha_{secmax}}{2}
$
for all $m \in \mathcal{M}$.
Then we have that
\[
\beta_{i, A(\bm{\alpha})}
>  \alpha_{A(\bm{\alpha})}
 - \frac{\alpha_{max} - \alpha_{secmax}}{2}
=  \frac{\alpha_{max} + \alpha_{secmax}}{2}.
\]
Consider $m = A(\bm{\alpha})$, we have
\begin{align*}
\beta_{i, m}
&
<  \alpha_{m}
+ \frac{\alpha_{max} - \alpha_{secmax}}{2}
\\
& 
\leq
\alpha_{secmax}
+ \frac{\alpha_{max} - \alpha_{secmax}}{2}
\\
&
=
  \frac{\alpha_{max} + \alpha_{secmax}}{2}.
\end{align*}
Namely, the true quality is still revealed.
\done
 
\end{document}